%% file: main.tex
\documentclass[acmsmall]{acmart}

\AtBeginDocument{%
  \providecommand\BibTeX{{%
    \normalfont B\kern-0.5em{\scshape i\kern-0.25em b}\kern-0.8em\TeX}}}

\newcommand{\paratitle}[1]{\textbf{#1}}
\newcommand{\term}[1]{\index{\lowercase{#1}}\emph{#1}}

\usepackage{multirow}
\usepackage{makecell}
\usepackage{makecell}
\usepackage{xspace,mfirstuc,tabulary}
\usepackage{booktabs}
\usepackage{graphicx}
\usepackage{bbding}
\usepackage{xcolor,colortbl}
\usepackage{multirow}
\usepackage{adjustbox}

\usepackage{pifont}
\newrobustcmd{\B}{\bfseries}
\newcommand*\colourcheck[1]{%
  \expandafter\newcommand\csname #1check\endcsname{\textcolor{#1}{\ding{52}}}%
}
\newcommand*\colourcross[1]{%
  \expandafter\newcommand\csname #1cross\endcsname{\textcolor{#1}{\ding{55}}}%
}
\colourcheck{green}
\colourcross{red}
\newcommand{\eg}{\emph{e.g.}}

\usepackage[most]{tcolorbox}
\newtcolorbox{highlighted}{
  colback=yellow!50!white, colframe=yellow!50!white, boxrule=0pt, sharp corners, left=0pt, right=0pt, top=0pt, bottom=0pt
}

\usepackage{color,soul}
\soulregister{\textcolor}{2}
\soulregister{\cite}7
\soulregister{\citep}7
\soulregister{\citet}7
\soulregister{\ref}7

\usepackage[switch]{lineno}
\sethlcolor{yellow}

\usepackage[edges]{forest}
\definecolor{lightcoral}{rgb}{0.94, 0.5, 0.5}
\definecolor{lightgreen}{rgb}{0.56, 0.93, 0.56}
\definecolor{harvestgold}{rgb}{0.98, 0.85, 0.40}
\definecolor{brightlavender}{rgb}{0.75, 0.58, 0.89}
\definecolor{capri}{rgb}{0.0, 0.75, 1.0}
\definecolor{carminepink}{rgb}{0.92, 0.3, 0.26}
\definecolor{celadon}{rgb}{0.67, 0.88, 0.69}
\definecolor{darkpastelgreen}{rgb}{0.01, 0.75, 0.24}

\definecolor{hidden-draw}{RGB}{205, 44, 36}
\definecolor{hidden-blue}{RGB}{194,232,247}
\definecolor{hidden-orange}{RGB}{243,202,120}
\definecolor{hidden-yellow}{RGB}{242,244,193}
\definecolor{tree-level-1}{RGB}{245,20,85}
\definecolor{tree-level-2}{RGB}{246,86,118}
\definecolor{tree-level-3}{RGB}{248,177,193}
\definecolor{tree-leaf}{RGB}{176,230,198}
\setcopyright{acmlicensed}
\acmJournal{TOIS}
\acmYear{2024}
\acmVolume{1}
\acmNumber{1}
\acmArticle{1}
\acmMonth{1}
\acmDOI{10.1145/3703155}

\begin{document}

\title{A Survey on Hallucination in Large Language Models: Principles, Taxonomy, Challenges, and Open Questions}

\author{Lei Huang}
\affiliation{%
	\institution{Harbin Institute of Technology}
	\streetaddress{800 Dongchuan Road}
	\city{Harbin}
	\state{Heilongjiang}
	\country{China}
	\postcode{150001}}
\email{lhuang@ir.hit.edu.cn}

\author{Weijiang Yu}
\affiliation{%
	\institution{Huawei Inc.}
	\streetaddress{Bantian Subdistrict}
	\city{Shenzhen}
	\state{Guangdong}
	\country{China}
	\postcode{518129}}
\email{weijiangyu8@gmail.com}

\author{Weitao Ma}
\email{wtma@ir.hit.edu.cn}
\author{Weihong Zhong}
\email{whzhong@ir.hit.edu.cn}
\affiliation{%
	\institution{Harbin Institute of Technology}
	\streetaddress{800 Dongchuan Road}
	\city{Harbin}
	\state{Heilongjiang}
	\country{China}
	\postcode{150001}}

\author{Zhangyin Feng}
\email{zyfeng@ir.hit.edu.cn}
\author{Haotian Wang}
\email{wanght1998@gmail.com}
\affiliation{%
	\institution{Harbin Institute of Technology}
	\streetaddress{800 Dongchuan Road}
	\city{Harbin}
	\state{Heilongjiang}
	\country{China}
	\postcode{150001}}

\author{Qianglong Chen}
\email{chenqianglong.ai@gmail.com}
\author{Weihua Peng}
\email{pengwh.hit@gmail.com}
\affiliation{%
	\institution{Huawei Inc.}
	\streetaddress{Bantian Subdistrict}
	\city{Shenzhen}
	\state{Guangdong}
	\country{China}
	\postcode{518129}}

\author{Xiaocheng Feng}\authornote{corresponding author}
\email{xcfeng@ir.hit.edu.cn}
\author{Bing Qin}
\email{qinb@ir.hit.edu.cn}
\author{Ting Liu}
\email{tliu@ir.hit.edu.cn}
\affiliation{%
	\institution{Harbin Institute of Technology}
	\streetaddress{800 Dongchuan Road}
	\city{Harbin}
	\state{Heilongjiang}
	\country{China}
	\postcode{150001}}
\renewcommand{\shortauthors}{Huang, et al.}

\begin{abstract}
  The emergence of large language models (LLMs) has marked a significant breakthrough in natural language processing (NLP), fueling a paradigm shift in information acquisition. Nevertheless, LLMs are prone to hallucination, generating plausible yet nonfactual content. This phenomenon raises significant concerns over the reliability of LLMs in real-world information retrieval (IR) systems and has attracted intensive research to detect and mitigate such hallucinations. Given the open-ended general-purpose attributes inherent to LLMs, LLM hallucinations present distinct challenges that diverge from prior task-specific models. This divergence highlights the urgency for a nuanced understanding and comprehensive overview of recent advances in LLM hallucinations. In this survey, we begin with an innovative taxonomy of hallucination in the era of LLM and then delve into the factors contributing to hallucinations. Subsequently, we present a thorough overview of hallucination detection methods and benchmarks. Our discussion then transfers to representative methodologies for mitigating LLM hallucinations. Additionally, we delve into the current limitations faced by retrieval-augmented LLMs in combating hallucinations, offering insights for developing more robust IR systems. Finally, we highlight the promising research directions on LLM hallucinations, including hallucination in large vision-language models and understanding of knowledge boundaries in LLM hallucinations. 
\end{abstract}




\maketitle

\input{1_Introduction}

\input{2_Definition}
\input{3_Causes}
\input{4_Benchmark_and_Detection}
\input{5_Mitigating}
\input{6_RAG}
\input{7_Future}
\input{8_Conclusion}

\bibliographystyle{ACM-Reference-Format}
\bibliography{custom}


\end{document}

%% file: 1_Introduction.tex
\section{Introduction}

Recently, the emergence of large language models (LLMs) \citep{zhao2023survey}, exemplified by LLaMA \citep{touvronllama1, touvron2023llama}, Claude \citep{2024clude}, Gemini \citep{anil2023gemini, reid2024gemini} and GPT-4 \citep{DBLP:journals/corr/abs-2303-08774}, has ushered in a significant paradigm shift in natural language processing (NLP), achieving unprecedented progress in language understanding \citep{hendrycks2020measuring, huang2023c}, generation \citep{zhang2023benchmarking, zhu2023multilingual} and reasoning \citep{wei2022chain, kojima2022large, qiao2022reasoning, yu2023nature, chu2023survey}. 
Furthermore, the extensive factual knowledge encoded within LLMs has demonstrated considerable advancements in leveraging LLMs for information seeking \citep{petroni2019language, alkhamissi2022a}, potentially reshaping the landscape of information retrieval systems \citep{zhu2023large}.
Nevertheless, in tandem with these remarkable advancements, concerns have arisen about the tendency of LLMs to generate hallucinations \citep{bang2023multitask, guerreiro2023hallucinations}, resulting in seemingly plausible yet factually unsupported content. 
Further compounding this issue is the capability of LLMs to generate highly convincing and human-like responses \citep{sadasivan2023can}, which makes detecting these hallucinations particularly challenging, thereby complicating the practical deployment of LLMs, especially real-world information retrieval (IR) systems that have integrated into our daily lives like chatbots \citep{2022chatgpt, 2023claude}, search engines \citep{2023bing, 2023perplexity}, and recommender systems \citep{gao2023chat, li2023gpt4rec}. Given that the information provided by these systems can directly influence decision-making, any misleading information has the potential to spread false beliefs, or even cause harm.

Notably, hallucinations in conventional natural language generation (NLG) tasks have been extensively studied \citep{huang2021the, ji2023survey}, with hallucinations defined as generated content that is either nonsensical or unfaithful to the provided source content. These hallucinations are categorized into two types: \textit{intrinsic hallucination}, where the generated output contradicts the source content, and \textit{extrinsic hallucination}, where the generated output cannot be verified from the source.
However, given their remarkable versatility across tasks \citep{bubeck2023sparks, bang2023multitask}, understanding hallucinations in LLMs presents a unique challenge compared to models tailored for specific tasks. Besides, as LLMs typically function as open-ended systems, the scope of hallucination encompasses a broader concept, predominantly manifesting factual errors.
This shift necessitates a reevaluation and adjustment of the existing taxonomy of hallucinations, aiming to enhance its adaptability in the evolving landscape of LLMs.

In this survey, we propose a redefined taxonomy of hallucination tailored specifically for applications involving LLMs. We categorize hallucination into two primary types: \textit{factuality hallucination} and \textit{faithfulness hallucination}. \textit{Factuality hallucination} emphasizes the discrepancy between generated content and verifiable real-world facts, typically manifesting as factual inconsistencies. 
Conversely, \textit{faithfulness hallucination} captures the divergence of generated content from user input or the lack of self-consistency within the generated content. This category is further subdivided into instruction inconsistency, where the content deviates from the user's original instruction; context inconsistency, highlighting discrepancies from the provided context; and logical inconsistency, pointing out internal contradictions within the content. Such categorization refines our understanding of hallucinations in LLMs, aligning it closely with their contemporary usage.

Delving into the underlying causes of hallucinations in LLMs is essential not merely for enhancing the comprehension of these phenomena but also for informing strategies aimed at alleviating them.
Recognizing the multifaceted sources of LLM hallucinations, our survey identifies potential contributors into three main aspects: data, training, and inference stages. This categorization allows us to span a broad spectrum of factors, providing a holistic view of the origins and mechanisms by which hallucinations may arise within LLM systems.
Furthermore, we comprehensively outline a variety of effective detection methods specifically devised for detecting hallucinations in LLMs, as well as an exhaustive overview of benchmarks related to LLM hallucinations, serving as appropriate testbeds to assess the extent of hallucinations generated by LLMs and the efficacy of detection methods. 
Beyond evaluation, significant efforts have been undertaken to mitigate hallucinations of LLMs. These initiatives are comprehensively surveyed in our study, in accordance with the corresponding causes, spanning from data-related, training-related, and inference-related approaches. 
In addition, the effectiveness of retrieval-augmented generation (RAG) in mitigating hallucinations has garnered tremendous attention within the field. Despite the considerable potential of RAG, current systems inherently face limitations and even suffer from hallucinations. Accordingly, our survey undertakes an in-depth analysis of these challenges, aiming to provide valuable insights aimed at developing more robust RAG systems. We also highlight several promising avenues for future research, such as hallucinations in large vision-language models and understanding of knowledge boundaries in LLM hallucinations, paving the way for forthcoming research in the field.

\paratitle{\textbf{Comparing with Existing Surveys.}}
As hallucination stands out as a major challenge in generative AI, numerous research \citep{ji2023survey, rawte2023survey, liu2023trustworthy, zhang2023siren, wang2023survey, tonmoy2024a} has been directed towards hallucinations.
While these contributions have explored LLM hallucination from various perspectives and provided valuable insights, our survey seeks to delineate their distinct contributions and the comprehensive scope they encompass.
\citet{ji2023survey} primarily shed light on hallucinations in pre-trained models for NLG tasks, leaving LLMs outside their discussion purview. 
\citet{tonmoy2024a} mainly focused on discussing the mitigation strategies combating LLM hallucinations.
Besides, \citet{liu2023trustworthy} took a broader view of LLM trustworthiness without delving into specific hallucination phenomena, whereas \citet{wang2023survey} provided an in-depth look at factuality in LLMs. 
However, our work narrows down to a critical subset of trustworthiness challenges, specifically addressing factuality and extending the discussion to include faithfulness hallucinations. 
To the best of our knowledge, \citet{zhang2023siren} presented research closely aligned with ours, detailing LLM hallucination taxonomies, evaluation benchmarks, and mitigation strategies. However, our survey sets itself apart through a unique taxonomy and organizational structure. We present a detailed, layered classification of hallucinations and conduct a more comprehensive analysis of the causes of hallucinations. Crucially, our proposed mitigation strategies are directly tied to these causes, offering a targeted and coherent framework for addressing LLM hallucinations.
\input{figures/category_tree_new.tex}

\paratitle{\textbf{Organization of this Survey.}} In this survey, we present a comprehensive overview of the latest developments in LLM hallucinations, as shown in Fig~\ref{categorization_of_survey}. We commence by constructing a taxonomy of hallucinations in the realm of LLM (\S\ref{sec:definition}). Subsequently, we analyze factors contributing to LLM hallucinations in depth (\S\ref{sec:causes}), followed by a review of various strategies and benchmarks employed for the reliable detection of hallucinations in LLMs (\S\ref{sec:detection_and_benchmark}). We then detail a spectrum of approaches designed to mitigate these hallucinations (\S\ref{sec:mitigating}). Concluding, we delve into the challenges faced by current RAG systems (\S\ref{sec:hallucination_rag}) and delineate potential pathways for forthcoming research (\S\ref{sec:future}).

%% file: figures/category_tree_new.tex
\tikzstyle{my-box}=[
    rectangle,
    draw=hidden-draw,
    rounded corners,
    text opacity=1,
    minimum height=1.5em,
    minimum width=5em,
    inner sep=2pt,
    align=center,
    fill opacity=.5,
]
\tikzstyle{cause_leaf}=[my-box, minimum height=1.5em,
    fill=harvestgold!20, text=black, align=left,font=\scriptsize,
    inner xsep=2pt,
    inner ysep=4pt,
]
\tikzstyle{detect_leaf}=[my-box, minimum height=1.5em,
    fill=cyan!20, text=black, align=left,font=\scriptsize,
    inner xsep=2pt,
    inner ysep=4pt,
]
\tikzstyle{mitigate_leaf}=[my-box, minimum height=1.5em,
    fill=lightgreen!20, text=black, align=left,font=\scriptsize,
    inner xsep=2pt,
    inner ysep=4pt,
]
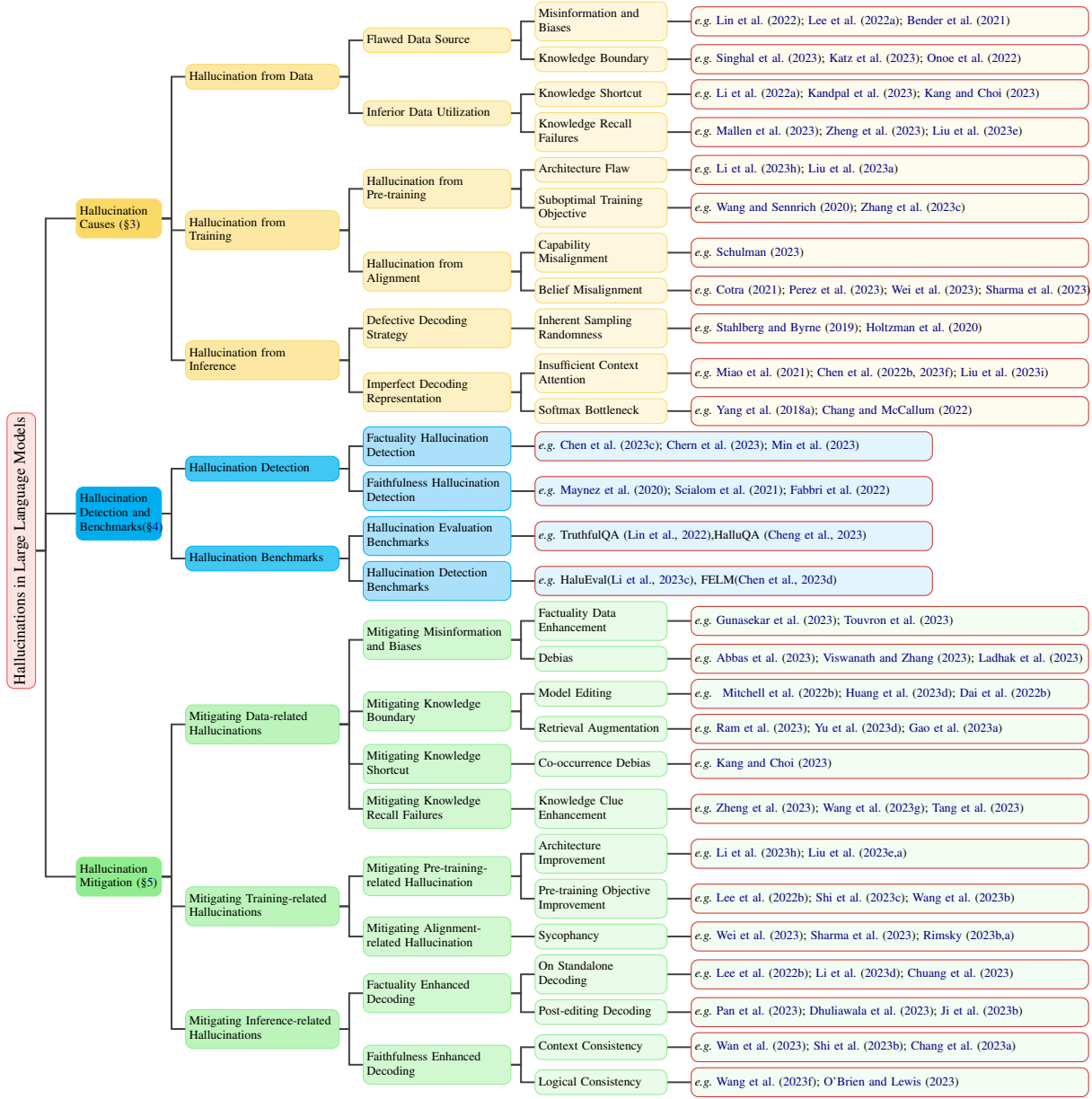
\begin{figure*}[tp]
    \centering
    \resizebox{\textwidth}{!}{
        \begin{forest}
            forked edges,
            for tree={
                grow=east,
                reversed=true,
                anchor=base west,
                parent anchor=east,
                child anchor=west,
                base=left,
                font=\small,
                rectangle,
                draw=hidden-draw,
                rounded corners,
                align=left,
                minimum width=4em,
                edge+={darkgray, line width=1pt},
                s sep=3pt,
                inner xsep=2pt,
                inner ysep=3pt,
                ver/.style={rotate=90, child anchor=north, parent anchor=south, anchor=center},
            },
            where level=1{text width=4.5em,font=\scriptsize,}{},
            where level=2{text width=8em,font=\scriptsize,}{},
            where level=3{text width=7.5em,font=\scriptsize,}{},
            where level=4{text width=7em,font=\scriptsize,}{},
            [
                Hallucinations in Large Language Models, ver, color=carminepink!100, fill=carminepink!15,
                text=black
                [
                    Hallucination \\ Causes (\S \ref{sec:causes}), color=harvestgold!100, fill=harvestgold!100, text=black
                    [
                        Hallucination from Data, color=harvestgold!100, fill=harvestgold!60,  text=black
                        [
                            Misinformation and \\ Biases, color=harvestgold!100, fill=harvestgold!40, text=black
                            		[
                                		{\eg~\citet{DBLP:conf/acl/LinHE22, lee2021deduplicating, bender2021dangers}}
                                		, cause_leaf, text width=20em
                            		]
                        ]
                        [
                            Knowledge Boundary, color=harvestgold!100, fill=harvestgold!40, text=black
                            		[
                                		{\eg~\citet{singhal2023towards, katz2023gpt, onoe2022entity}}
                                		, cause_leaf, text width=20em
                            		]
                        ]
                        [
                            Inferior Alignment Data, color=harvestgold!100, fill=harvestgold!40, text=black
                            		[
                                		{\eg~\citet{gekhman2024does, li2024the}}
                                		, cause_leaf, text width=20em
                            		]
                        ]
                    ]
                    [
                        Hallucination from \\ Training, color=harvestgold!100, fill=harvestgold!60, text=black
                        [
                            Hallucination from \\ Pre-training, color=harvestgold!100, fill=harvestgold!40, text=black
                            		[
                                		{\eg~\citet{li2023batgpt, liu2023exposing, wang2020exposure}}
                                		, cause_leaf, text width=20em
                            		]
                        ]
                        [
                            Hallucination from SFT, color=harvestgold!100, fill=harvestgold!40, text=black
                            		[
                                		{\eg~\citet{schulman2023youtube, yang2023alignment, zhang2023rtuning}}
                                		, cause_leaf, text width=20em
                            		]
                        ]
                         [
                            Hallucination from RLHF, color=harvestgold!100, fill=harvestgold!40, text=black
                            		[
                                		{\eg~\citet{cotra2021why, perez2022discovering, wei2023simple, sharma2023towards}}
                                		, cause_leaf, text width=20em
                            		]
                        ]
                    ]
                    [
                        Hallucination from \\ Inference, color=harvestgold!100, fill=harvestgold!60, text=black
                        [
                            Imperfect Decoding \\ Strategies, color=harvestgold!100, fill=harvestgold!40, text=black
                            [
                                {\eg~\citet{stahlberg2019nmt, holtzman2019curious}}
                                , cause_leaf, text width=20em
                            ]
                        ]
                        [
                            Over-confidence, color=harvestgold!100, fill=harvestgold!40, text=black
                            		[
                                		{\eg~\citet{miao2021prevent, chen2022towards, chen2023improving, liu2023instruction}}
                                		, cause_leaf, text width=20em
                            		]
                         ]
                         [
                            Softmax Bottleneck, color=harvestgold!100, fill=harvestgold!40, text=black
                            		[
                                		{\eg~\citet{miao2021prevent, chang2022softmax}}
                                		, cause_leaf, text width=20em
                            		]
                         ]     
                         [
                            Reasoning Failure, color=harvestgold!100, fill=harvestgold!40, text=black
                            		[
                                		{\eg~\citet{zheng2023does, berglund2023reversal}}
                                		, cause_leaf, text width=20em
                            		]
                         ]                    
                    ] 
                ]
                [
                    Hallucination \\ Detection and \\ Benchmarks(\S \ref{sec:detection_and_benchmark}), color=cyan!100, fill=cyan!100, text=black
                    [
                        Hallucination Detection, color=cyan!100, fill=cyan!60, text=black
                        [
                            Factuality Hallucination \\ Detection, color=cyan!100, fill=cyan!30, text=black
                            [
                                {\eg~\citet{DBLP:journals/corr/abs-2305-14251, dhuliawala2023chain, DBLP:journals/corr/abs-2303-08896}}
                                , detect_leaf, text width=20em
                            ]
                        ]
                        [
                            Faithfulness Hallucination \\ Detection, color=cyan!100, fill=cyan!30, text=black
                            [
                                {\eg~\citet{maynez2020faithfulness, scialom2021questeval, fabbri2021qafacteval}}
                                , detect_leaf, text width=20em
                            ]
                        ]
                    ]
                    [
                        Hallucination Benchmarks, color=cyan!100, fill=cyan!60, text=black
                        [
                            Hallucination Evaluation \\ Benchmarks, color=cyan!100, fill=cyan!30, text=black
                            [
                                {\eg~TruthfulQA \citep{DBLP:conf/acl/LinHE22}, HalluQA \citep{cheng2023evaluating}, HaluEval-2.0 \citep{li2024the}}
                                , detect_leaf, text width=20em
                            ]
                        ]
                        [
                            Hallucination Detection \\ Benchmarks, color=cyan!100, fill=cyan!30, text=black
                            [
                                {\eg~SelfCheckGPT-Wikibio~\citep{miao2023selfcheck}, HaluEval~\citep{li2023halueval}, FELM~\citep{chen2023felm}}
                                , detect_leaf, text width=20em
                            ]
                        ]
                    ]
                ]
                [
                    Hallucination \\ Mitigation (\S \ref{sec:mitigating}), color=lightgreen!100, fill=lightgreen!100, text=black
                    [
                        Mitigating Data-related \\ Hallucinations, color=lightgreen!100, fill=lightgreen!60, text=black
                        [
                            Data Filtering, color=lightgreen!100, fill=lightgreen!40, text=black
                            		[
                                		{\eg~\citet{gunasekar2023textbooks, touvron2023llama, abbas2023semdedup}}
                                , mitigate_leaf, text width=20em
                            		]
                        ]
                        [
                            Model Editing, color=lightgreen!100, fill=lightgreen!40, text=black
                            		[
                                		{\eg~ \citet{mitchell2022memory, huang2023transformer, dai2022neural}}
                                , mitigate_leaf, text width=20em
                                ]
                         ]
                          [
                            		Retrieval-Augmented \\ Generation, color=lightgreen!100, fill=lightgreen!40, text=black
                            		[
                                		{\eg~\citet{DBLP:journals/corr/abs-2302-00083, yu2023improving, DBLP:conf/acl/GaoDPCCFZLLJG23}}
                                , mitigate_leaf, text width=20em
                            		]
                            ]
                        ]
                    [
                        Mitigating Training-related \\ Hallucinations, color=lightgreen!100, fill=lightgreen!60, text=black
                        [
                            Mitigating Pre-training-\\related Hallucination, color=lightgreen!100, fill=lightgreen!40, text=black
                            [
                                {\eg~\citet{li2023batgpt, DBLP:journals/corr/abs-2307-03172, liu2023exposing, shi2023context}}
                                , mitigate_leaf, text width=20em
                            ]
                        ]
                        [
                            Mitigating Misalignment \\ Hallucination, color=lightgreen!100, fill=lightgreen!40, text=black
                            [
                                {\eg~\citet{wei2023simple, sharma2023towards, rimsky2023reducing}}
                                , mitigate_leaf, text width=20em
                            ]
                        ]
                    ]
                    [
                        Mitigating Inference-related \\ Hallucinations, color=lightgreen!100, fill=lightgreen!60, text=black
                        [
                            Factuality Enhanced \\ Decoding, color=lightgreen!100, fill=lightgreen!40, text=black
                            [
                                {\eg~\citet{lee2022factuality, li2023inference, chuang2023dola}}
                                , mitigate_leaf, text width=20em
                            ]
                        ]
                        [
                            Faithfulness Enhanced \\ Decoding, color=lightgreen!100, fill=lightgreen!40, text=black
                            [
                                {\eg~\citet{wan2023faithfulness, DBLP:journals/corr/abs-2305-14739, DBLP:journals/corr/abs-2306-01286}}
                                , mitigate_leaf, text width=20em
                            ]
                        ]
                   ]
               ]
        ]
        \end{forest}
    }
    \caption{The main content flow and categorization of this survey.}
    \label{categorization_of_survey}
\end{figure*}

%% file: 2_Definition.tex
\section{Definitions}
\label{sec:definition}

For the sake of a comprehensive understanding of hallucinations in LLMs, we commence with a succinct introduction to LLMs (\S\ref{ssec:large_language_model}), delineating the scope of this survey. Subsequently, we delve into the training stages of LLMs (\S\ref{ssec:training_llm}), as a thorough understanding of the training mechanisms contributes significantly to elucidating the origins of hallucinations. Lastly, we expound upon the concept of hallucinations in LLMs (\S\ref{ssec:llm_hallucination}), further categorizing it into two distinct types.

\subsection{Large Language Models}
\label{ssec:large_language_model}

Before delving into the causes of hallucination, we first introduce the concept of LLMs. Typically, LLMs refer to a series of general-purpose models that leverage the transformer-based language model architecture and undergo extensive training on massive textual corpora with notable examples including GPT-3 \citep{brown2020language}, PaLM \citep{chowdhery2022palm}, LLaMA \citep{touvron2023llama}, GPT-4 \citep{DBLP:journals/corr/abs-2303-08774} and Gemini \citep{reid2024gemini}. By scaling the amount of data and model capacity, LLMs raise amazing emergent abilities, typically including in-context learning (ICL) \citep{brown2020language}, chain-of-thought prompting \citep{wei2022chain} and instruction following \citep{peng2023instruction}.

\subsection{Training Stages of Large Language Models}
\label{ssec:training_llm}

The attributes and behaviors of LLMs are deeply intertwined with their training processes. LLMs undergo three primary training stages: pre-training, supervised fine-tuning (SFT), and reinforcement learning from human feedback (RLHF). Analyzing these stages provides insight into hallucination origins in LLMs, as each stage equips the model with specific capabilities.

\subsubsection{Pre-training}
Pre-training is widely acknowledged as a foundational stage for LLM to acquire knowledge and capabilities \citep{zhou2023lima}. During this phase, LLMs engage in autoregressive prediction of subsequent tokens within sequences. Through self-supervised training on extensive textual corpora, LLMs acquire knowledge of language syntax, world knowledge, and reasoning abilities, thereby laying a solid groundwork for further fine-tuning.
Besides, recent research \citep{youtube, deletang2023language} suggests that predicting subsequent words is akin to losslessly compressing significant information. The essence of LLMs lies in predicting the probability distribution for upcoming words. Accurate predictions indicate a profound grasp of knowledge, translating to a nuanced understanding of the world.

\subsubsection{Supervised Fine-Tuning}
While LLMs acquire substantial knowledge and capabilities during the pre-training stage, it's crucial to recognize that pre-training primarily optimizes for completion. Consequently, pre-trained LLMs fundamentally serve as completion machines, which can lead to a misalignment between the next-word prediction objective of LLMs and the user's objective of obtaining desired responses. To bridge this gap, SFT \citep{zhang2023instruction} has been introduced, which involves further training LLMs using a meticulously annotated set of (instruction, response) pairs, resulting in enhanced capabilities and improved controllability of LLMs. Furthermore, recent studies \citep{chung2022scaling, iyer2022opt} have confirmed the effectiveness of supervised fine-tuning to achieve exceptional performance on unseen tasks, showcasing their remarkable generalization abilities.

\subsubsection{Reinforcement Learning from Human Feedback}
While the SFT process successfully enables LLMs to follow user instructions, there is still room for them to better align with human preferences. 
Among various methods that utilize human feedback, RLHF stands out as an representative solution for aligning with human preferences through reinforcement learning \citep{christiano2017deep, stiennon2020learning,ouyang2022training}. 
Typically, RLHF employs a preference model \citep{bradley1952rank} trained to predict preference rankings given a prompt alongside a pair of human-labeled responses. 
To align with human preferences, RLHF optimizes the LLM to generate outputs that maximize the reward provided by the trained preference model, typically employing a reinforcement learning algorithm, such as Proximal Policy Optimization (PPO) \citep{schulman2017proximal}. 
Such integration of human feedback into the training loop has proven effective in enhancing the alignment of LLMs, guiding them toward producing high-quality and harmless responses.

\input{tables/new_example}

\subsection{Hallucinations in Large Language Models} 
\label{ssec:llm_hallucination}

The concept of hallucination traces its roots to the fields of pathology and psychology and is defined as \textit{the perception of an entity or event that is absent in reality} \citep{macpherson2013hallucination}. 
Within the realm of NLP, hallucination is typically referred to as a phenomenon in which the generated content appears nonsensical or unfaithful to the provided source content \citep{filippova2020controlled, maynez2020faithfulness}. This concept bears a loose resemblance to the phenomenon of hallucination observed in human psychology. 
Generally, hallucinations in natural language generation tasks can be categorized into two primary types: \textit{intrinsic hallucination} and \textit{extrinsic hallucination} \citep{huang2021factual, li2022faithfulness, ji2023survey}. 

Specifically, \textit{intrinsic hallucinations} pertain to the model outputs that directly conflict with the provided source context. 
On the other hand, \textit{extrinsic hallucinations} involve outputs that cannot be verified using the provided source context or external knowledge bases. This means the generated text is neither supported by nor directly contradicts the available information, rendering the output unverifiable and potentially misleading.

However, in the era of LLMs, the versatile capabilities of these models have facilitated their widespread use across diverse fields, highlighting limitations in existing task-specific categorization paradigms. Considering that LLMs place a significant emphasis on user-centric interactions and prioritize alignment with user directives, coupled with the fact that their hallucinations predominantly surface at factual levels, we introduce a more granular taxonomy building upon the foundational work by \citet{ji2023survey}. This refined taxonomy seeks to encapsulate the distinct intricacies associated with LLM hallucinations. To provide a more intuitive illustration of our definition of LLM hallucination, we present examples for each type of hallucination in Table \ref{tab:hallucination_example}, namely \textit{factuality hallucination} and \textit{faithfulness hallucination}.

\subsubsection{Factuality Hallucination}
\label{sssec:factuality hallucination}
The emergence of LLMs marks a significant shift from traditional task-specific toolkits to AI assistants that have a heightened focus on open-domain interactions. This shift is primarily attributed to their vast parametric factual knowledge. However, existing LLMs occasionally exhibit tendencies to produce outputs that are either inconsistent with real-world facts or unverifiable \citep{li2024the}, posing challenges to the trustworthiness of artificial intelligence. In this context, we categorize these factuality hallucinations into two primary types:

\paratitle{Factual Contradiction} refers to situations where the LLM's output contains facts that can be grounded in real-world information, but present contradictions. This type of hallucination occurs most frequently and arises from diverse sources, encompassing the LLM's capture, storage, and expression of factual knowledge. Depending on the error type of contradictions, it can be further divided into two subcategories: \textit{entity-error hallucination} and \textit{relation-error hallucination}.
\begin{itemize}
	\item \textbf{Entity-error hallucination} refers to the situations where the generated text of LLMs contains erroneous entities. As shown in Table \ref{tab:hallucination_example}, when asked about \textit{"the inventor of the telephone"}, the model erroneously states \textit{"Thomas Edison"}, conflicting with the real fact that it was \textit{"Alexander Graham Bell"}.
	\item \textbf{Relation-error hallucination} refers to instances where the generated text of LLMs contains wrong relations between entities. As shown in Table 1, when inquired about \textit{"the inventor of the light bulb"}, the model incorrectly claims \textit{"Thomas Edison"}, despite the fact that \textit{he improved upon existing designs and did not invent it}.
\end{itemize}

\paratitle{Factual Fabrication} refers to instances where the LLM's output contains facts that are unverifiable against established real-world knowledge. This can be further divided into \textit{unverifiability hallucination} and \textit{overclaim hallucination}.
\begin{itemize}
	\item \textbf{Unverifiability hallucination} pertains to statements that are entirely non-existent or cannot be verified using available sources. As shown in Table \ref{tab:hallucination_example}, when asked about \textit{"the major environmental impacts of the construction of the Eiffel Tower"}, the model incorrectly states that \textit{"the construction led to the extinction of the Parisian tiger"}, a species that does not exist and thus, this claim cannot be substantiated by any historical or biological record.
	\item \textbf{Overclaim hallucination} involves claims that lack universal validity due to subjective biases. As shown in Table 1, the model claims that \textit{"the Eiffel Tower's construction is widely recognized as the event that sparked the global green architecture movement."} This is an overclaim, as there is no broad consensus or substantial evidence to support the statement.
\end{itemize}
\subsubsection{Faithfulness Hallucination}
\label{sssec:faithfulness hallucination}
LLMs are inherently trained to align with user instructions. As the use of LLMs shifts towards more user-centric applications, ensuring their consistency with user-provided instructions and contextual information becomes increasingly vital. Furthermore,  LLM's faithfulness is also reflected in the logical consistency of its generated content. From this perspective, we categorize three subtypes of faithfulness hallucinations:

\paratitle{Instruction inconsistency} refers to the LLM's outputs that deviate from a user's directive. While some deviations might serve safety guidelines, the inconsistencies here signify unintentional misalignment with non-malicious user instructions. As described in Table \ref{tab:hallucination_example}, the user's actual intention is translation, However, the LLM erroneously deviated from the user's instruction and performed a question-answering task instead.

\paratitle{Context inconsistency} points to instances where the LLM's output is unfaithful with the user's provided contextual information. For example, as shown in Table \ref{tab:hallucination_example}, the user mentioned the Nile's source being in the Great Lakes region of central Africa, yet the LLM's response contradicted the context.

\paratitle{Logical inconsistency} underscores when LLM outputs exhibit internal logical contradictions, often observed in reasoning tasks. This manifests as inconsistency both among the reasoning steps themselves and between the steps and the final answer. For example, as shown in Table \ref{tab:hallucination_example}, while the reasoning step of dividing both sides of the equation by 2 is correct, the final answer of x=4 is inconsistent with the reasoning chain, leading to an incorrect result.

%% file: tables/new_example.tex
\begin{table*}[]
\centering
\caption{Examples of each category of LLM hallucinations. Content marked in \textcolor{red}{Red} represents the hallucinatory output, while content marked in \textcolor{blue}{Blue} indicates user instruction or provided context that contradicts the LLM hallucination.}
  \label{tab:hallucination_example}.
\label{tab:causes}
\vspace{-4mm}  
\begin{adjustbox}{max width=\textwidth}\setlength{\tabcolsep}{0.3mm}
{
\begin{tabular}{llll}
\toprule
\textbf{Category} & \textbf{Type} & \textbf{Example Dialog} & \textbf{Explanation} 
\\\midrule
\multicolumn{1}{c}{} & \begin{tabular}[c]{@{}l@{}} Factual \\Contradiction \end{tabular} &
  \begin{tabular}[c]{@{}l@{}}
  \textbf{Instruction:} What are Thomas Edison's main contributions \\to science and technology?\\
   \midrule
\textbf{Response: }Thomas Edison developed \textcolor{red}{the first practical tele-}\\\textcolor{red}{phone} and \textcolor{red}{invented} the light bulb.
  \\
  \end{tabular} & 
    \begin{tabular}[c]{@{}l@{}}
The response is factually incorrect. In reality, \textbf{Edison}\\ \textbf{improved the light bulb, building on earlier desi-} \\ \textbf{gns} \textbf{by others}, and \textbf{Alexander Graham Bell inven}-\\ \textbf{ted the telephone}. 
\end{tabular} 
   \\ \cmidrule(lr){2-4}
\multicolumn{1}{c}{\multirow{-6}{*}{\begin{tabular}[c]{@{}l@{}}Factuality\\Hallucination\end{tabular}}} & \begin{tabular}[c]{@{}l@{}} Factual \\ Fabrication \end{tabular} &
  \begin{tabular}[c]{@{}l@{}}
 \textbf{Instruction:} What are the major environmental impacts of \\the construction of the Eiffel Tower?\\

  \midrule
\textbf{Response:} The construction of the Eiffel Tower in 1889 led\\ to \textcolor{red}{the extinction of the Parisian tiger}, a species that played a \\crucial role in the region's ecosystem. Additionally, \textcolor{red}{it is wid-} \\\textcolor{red}{ely recognized as the event that sparked the global green} \\\textcolor{red}{architecture movement}. \end{tabular} & 
\begin{tabular}[c]{@{}l@{}} The response erroneously states that the construction\\ of the Eiffel Tower led to the extinction of the `Parisian\\ tiger' and sparked the global green architecture move- \\ment. In fact, \textbf{there is no evidence of a `Parisian tig-}\\ \textbf{er' ever existing, making this a fabricated claim.}\\ Moreover, \textbf{attributing the origins of green architec-}\\ \textbf{ture to the Eiffel Tower is an exaggeration}, as this \\ movement has diverse roots and cannot be traced back \\to a single event.
   \end{tabular}
   \\ \midrule
\multicolumn{1}{c}{} &
  \begin{tabular}[c]{@{}l@{}} Instruction \\Inconsistency \end{tabular} &
  \begin{tabular}[c]{@{}l@{}}
    \textbf{Instruction:} \textcolor{blue}{Translate the English question into Spanish}: \\"What is the capital of France?"\\
    \midrule
    
    \textbf{Response:} \textcolor{red}{The capital of France is Paris.}
    \\ 
    \end{tabular} &
    \begin{tabular}[c]{@{}l@{}}
    The user explicitly provided an instruction to translate \\ the English question into Spanish. \textbf{Instead of perform-} \\ \textbf{ing the translation as instructed, the LLM chose to} \\ \textbf{answer the question.}\end{tabular} 
   \\ \cmidrule(lr){2-4}
\multicolumn{1}{c}{} &  \begin{tabular}[c]{@{}l@{}} Context \\ Inconsistency \end{tabular} &
  \begin{tabular}[c]{@{}l@{}}
  \textbf{Retrieved context:} The Nile is a major north-flowing river \\ in northeastern Africa. It's traditionally considered the \\ longest river in the world. \textcolor{blue}{Originating in the Great Lakes} \\ \textcolor{blue}{region of central Africa}, it flows through several countries \\ before emptying into the Mediterranean Sea.\\
   \midrule
  \textbf{Instruction:} Summarize the retrieval document above. \\\midrule  \textbf{Response:} The Nile originates from \textcolor{red}{the mountain ranges} \\ \textcolor{red}{of central Africa} and flows into the Mediterranean.
  \\
  \end{tabular} &
  \begin{tabular}[c]{@{}l@{}}
    The summary incorrectly states that the Nile originates\\ from the "mountain ranges of central Africa", while the \\ \textbf{paragraph mentions that the Nile originates in the} \\ \textbf{Great Lakes region of central Africa.}
    \end{tabular}
   \\ \cmidrule(lr){2-4}
\multicolumn{1}{c}{\multirow{-18}{*}{\begin{tabular}[c]{@{}l@{}}Faithfulness\\Hallucination\end{tabular}}} & \begin{tabular}[c]{@{}l@{}} Logical \\ Inconsistency \end{tabular} &
  \begin{tabular}[c]{@{}l@{}}
 \textbf{Instruction:} Solve the equation 2x + 3 = 11 step by step.\\

  \midrule
  \textbf{Response:} Step 1: Subtract 3 from both sides to get: 2x = 8.\\ Step 2: Divide both sides by 2 to get: \textcolor{red}{x = 3. So the result} \\ \textcolor{red}{of the equation is x=3.} \end{tabular} & 
 \begin{tabular}[c]{@{}l@{}} The first step correctly isolates the term with x to get\\ 2x = 8. However, \textbf{the next step inaccurately divides} \\ \textbf{8 by 2 to yield a result of x = 3}, which is inconsistent \\with the earlier reasoning.
   \end{tabular}

  \\\bottomrule
\end{tabular}
}
\end{adjustbox}
\end{table*}

%% file: 3_Causes.tex
\section{Hallucination Causes}
\label{sec:causes}
LLM hallucinations have multifaceted origins, spanning the entire spectrum of LLMs' capability acquisition process. In this section, we delve into the root causes of hallucinations in LLMs, primarily categorized into three key aspects: (1) \textit{Data} (\S\ref{ssec:cause_data}), (2) \textit{Training} (\S\ref{ssec:cause_training}), and (3) \textit{Inference} (\S\ref{ssec:cause_decoding}).

\subsection{Hallucination from Data}
\label{ssec:cause_data}
Data for training LLMs are comprised of two primary components: (1) pre-training data, through which LLMs acquire their general capabilities and factual knowledge \citep{zhou2023lima}, and (2) alignment data, which teach LLMs to follow user instructions and align with human preferences \citep{wang2023aligning}. Although these data constantly expand the capability boundaries of LLMs, they inadvertently become the principal contributors to LLM hallucinations. This primarily manifests in three aspects: the presence of misinformation and biases in the flawed pre-training data sources (\S\ref{sssec:cause_data_misinformation_biases}), the knowledge boundary inherently bounded by the scope of the pre-training data (\S\ref{sssec:cause_data_knowledge_boundary}), and the hallucinations induced by inferior alignment data (\S\ref{sssec:cause_data_aligment}).

\subsubsection{Misinformation and biases}
\label{sssec:cause_data_misinformation_biases}
Neural networks possess an intrinsic tendency to memorize training data \citep{carlini2021extracting}, and this memorization tendency grows with model size \citep{carlini2022quantifying, chowdhery2022palm}.
In general, the inherent memorization capability is a double-edged sword in the fight against hallucinations.
On the one hand, the capacities of LLMs to memorize suggests their potential to capture profound world knowledge.
On the other hand, it becomes problematic in the context of misinformation and biases present within pre-training data and may inadvertently be amplified, manifesting as \textit{imitative falsehood} \citep{DBLP:conf/acl/LinHE22} and the reinforcement of societal biases. 
For a more comprehensive understanding, detailed examples are presented in Table \ref{data-induced_hallucination}.

\input{tables/bias_example}

\paratitle{Imitative Falsehood.}
Misinformation such as fake news and unfounded rumors has been widely spread among social media platforms and gradually serves as a significant contributor to LLM hallucinations.
The increasing demand for large-scale corpora for pre-training necessitates the employment of heuristic data collection methods.
While facilitating the acquisition of extensive data, challenges arise in maintaining consistent data quality, which inevitably introduces such misinformation in pre-training data \citep{bender2021dangers, weidinger2021ethical}.
In this situation, LLMs will increase the likelihood of generating such false statements due to their remarkable memorization capabilities, leading to \textit{imitative falsehoods}.
The issue is further exacerbated by the fact that LLMs have drastically lowered the barriers to content creation, posing considerable risks to the trustworthiness of public discourse and internet ecosystems.

\paratitle{Societal Biases.}
In addition to misinformation, biases are also deeply rooted in social media platforms, showing up variously, like biased hiring, prejudiced news, and hate-spewing.
As a purely negative phenomenon, biases and hallucinations have obvious differences, mainly manifested as stereotypes spreading and social inequalities. 
However, certain biases are intrinsically tied to hallucinations, especially those related to gender \citep{paullada2021data} and nationality \citep{venkit2023nationality, ladhak2023pre}.
For instance, LLMs might associate the profession of nursing with females, even when gender isn't explicitly mentioned in the user-provided context, exemplifying context inconsistency as discussed in Section (\S\ref{ssec:llm_hallucination}). 
Such biases can be inadvertently acquired from internet-based texts, which are rife with diverse and biased viewpoints, and subsequently be propagated into the generated content \citep{ladhak2023pre}. 

\input{tables/knowledge_boundary.tex}
\subsubsection{Knowledge Boundary.}
\label{sssec:cause_data_knowledge_boundary}
While the vast pre-training corpora have empowered LLMs with extensive factual knowledge, they inherently possess knowledge boundaries. 
These boundaries arise primarily from two sources: (1) the inability of LLMs to memorize all factual knowledge encountered during pre-training, especially the less frequent long-tail knowledge; and (2) the intrinsic boundary of the pre-training data itself, which does not include rapidly evolving world knowledge or content restricted by copyright laws.
Consequently, when LLMs encounter information that falls outside their limited knowledge boundaries, they are more susceptible to generating hallucinations.
We present detailed examples for clear illustration in Table \ref{tab:knowledge_boundary}.

\paratitle{Long-tail Knowledge.}
The distribution of knowledge within the pre-training corpora is inherently non-uniform, which results in LLMs demonstrating varying levels of proficiency across different types of knowledge.
Recent studies have highlighted a strong correlation between the model's accuracy on general domain questions and the volume of relevant documents \citep{kandpal2023large} or entity popularity \citep{DBLP:conf/acl/MallenAZDKH23} within the pre-training corpora.
Furthermore, given that LLMs are predominantly trained on extensive general domain corpora \citep{penedo2023refinedweb, raffel2020exploring, gao2020pile}, they may exhibit deficits in domain-specific knowledge.
This limitation becomes particularly evident when LLMs are confronted with tasks that require domain-specific expertise, such as medical \citep{li2023chatdoctor, singhal2023towards} and legal \citep{yu2022legal, katz2023gpt} questions, these models may exhibit pronounced hallucinations, often manifesting as factual fabrication.

\paratitle{Up-to-date Knowledge.}
Beyond the shortfall in long-tail knowledge, another intrinsic limitation concerning the knowledge boundaries within LLMs is their constrained capacity for up-to-date knowledge. The factual knowledge embedded within LLMs exhibits clear temporal boundaries and can become outdated over time \citep{onoe2022entity, kasai2022realtime, DBLP:conf/acl/LiRZWLVYK23}. Once these models are trained, their internal knowledge is never updated. This poses a challenge given the dynamic and ever-evolving nature of our world. When confronted with queries that transcend their temporal scope, LLMs often resort to fabricating facts or providing answers that might have been correct in the past but are now outdated. 

\paratitle{Copyright-sensitive Knowledge.}
Due to licensing restrictions \citep{reuters2023us}, existing LLMs are legally constrained to training on corpora that are publicly licensed \citep{gao2020pile, together2023redpajama} or otherwise available for use without infringing copyright laws \citep{henderson2022pile, arxiv2023arxiv}. This limitation significantly impacts the breadth and diversity of knowledge that LLMs can legally acquire. A significant portion of valuable knowledge, encapsulated in copyrighted materials such as recent scientific research, proprietary data, and copyrighted literary works, remains inaccessible to LLMs. This exclusion creates a knowledge gap, leading to potential hallucinations when LLMs attempt to generate information in domains where their training data is inaccessible \citep{min2023silo}.

\subsubsection{Inferior Alignment Data}
\label{sssec:cause_data_aligment}
After the pre-training stage, LLMs have embedded substantial factual knowledge within their parameters, thereby establishing obvious knowledge boundaries. 
During the supervised fine-tuning (SFT) stage, LLMs are typically trained on instruction pairs labeled by human annotators, potentially introducing new factual knowledge that extends beyond the knowledge boundary established during pre-training.
\citet{gekhman2024does} analyzed the training dynamics of incorporating new factual knowledge during the SFT process and found that LLMs struggle to acquire such new knowledge effectively. Most importantly, they discovered a correlation between the acquisition of new knowledge through SFT and increased hallucinations, suggesting that introducing new factual knowledge encourages LLMs to hallucinate.
Additionally, \citet{li2024the} conducted extensive analysis on the effect of instructions in producing hallucinations. Findings indicated that task-specific instructions which primarily focus on task format learning, tend to yield a higher proportion of hallucinatory responses. Moreover, overly complex and diverse instructions also lead to increased hallucinations.

\subsection{Hallucination from Training}
\label{ssec:cause_training}
As detailed in Section \ref{ssec:training_llm}, the distinct stages of training impart various capabilities to LLMs, with pre-training focusing on acquiring general-purpose representations and world knowledge, and alignment enables LLMs to better align with user instructions and preferences.
While these stages are critical for equipping LLMs with remarkable capabilities, shortfalls in either stage can inadvertently pave the way for hallucinations.

\subsubsection{Hallucination from Pre-training}
\label{sssec:cause_pretraining}
Pre-training constitutes the foundational stage for LLMs, predominantly utilizing a transformer-based architecture following the paradigm established by GPT \citep{radford2018improving, radford2019language, brown2020language}, and further developed by OPT\citep{zhang2022opt}, Falcon \citep{penedo2023refinedweb}, and Llama-2 \citep{touvron2023llama}.
This stage employs a causal language modeling objective, where models learn to predict subsequent tokens solely based on preceding ones in a unidirectional, left-to-right manner. While facilitating efficient training, it inherently limits the ability to capture intricate contextual dependencies, potentially increasing risks for the emergence of hallucination \citep{li2023batgpt}. 
Moreover, recent research has exposed that LLMs can occasionally exhibit unpredictable reasoning hallucinations spanning both long-range and short-range dependencies, which potentially arise from the limitations of soft attention \citep{hahn2020theoretical, chiang2022overcoming}, where attention becomes diluted across positions as sequence length increases.
Notably, the phenomenon of exposure bias \citep{bengio2015scheduled, ranzato2015sequence} has been a longstanding and serious contribution to hallucinations, resulting from the disparity between training and inference in the auto-regressive generative model.
Such inconsistency can result in hallucinations \citep{wang2020exposure}, especially when an erroneous token generated by the model cascades errors throughout the subsequent sequence, akin to a snowball effect \citep{zhang2023language}.

\subsubsection{Hallucination from Supervised Fine-tuning}
\label{sssec:cause_sft}
LLMs have inherent capability boundaries established during pre-training.
SFT seeks to utilize instruction data and corresponding responses to unlock these pre-acquired abilities. 
However, challenges arise when the demands of annotated instructions exceed the model's pre-defined capability boundaries.
In such cases, LLMs are trained to fit responses beyond their actual knowledge boundaries.
As discussed in \S\ref{sssec:cause_data_aligment}, over-fitting on new factual knowledge encourages LLMs prone to fabricating content, amplifying the risk of hallucinations \citep{schulman2023youtube, gekhman2024does}.
Moreover, another significant reason lies in the models' inability to reject.
Traditional SFT methods typically force models to complete each response, without allowing them to accurately express uncertainty \citep{yang2023alignment, zhang2023rtuning}. 
Consequently, when faced with queries that exceed their knowledge boundaries, these models are more likely to fabricate content rather than reject it.
This misalignment of knowledge boundaries, coupled with the inability to express uncertainty, are critical factors that contribute to the occurrence of hallucinations during the SFT stage.

\subsubsection{Hallucination from RLHF}
\label{sssec:cause_rlhf}
Several studies \citep{burns2022discovering, DBLP:journals/corr/abs-2304-13734} have demonstrated that LLM's activations encapsulate an internal belief related to the truthfulness of its generated statements. Nevertheless, misalignment can occasionally arise between these internal beliefs and the generated outputs.
Even when LLMs are refined with human feedback \citep{ouyang2022training}, they can sometimes produce outputs that diverge from their internal beliefs. 
Such behaviors, termed as sycophancy \citep{cotra2021why}, underscore the model's inclination to appease human evaluators, often at the cost of truthfulness. Recent studies indicate that models trained via RLHF exhibit pronounced behaviors of pandering to user opinions. Such sycophantic behaviors are not restricted to ambiguous questions without definitive answers \citep{perez2022discovering}, like political stances, but can also arise when the model chooses a clearly incorrect answer, despite being aware of its inaccuracy \citep{wei2023simple}. Delving into this phenomenon, \citet{sharma2023towards} suggested that the root of sycophancy may lie in the training process of RLHF models. By further exploring the role of human preferences in this behavior, the research indicates that the tendency for sycophancy is likely driven by both humans and preference models showing a bias towards sycophantic responses over truthful ones. 

\subsection{Hallucination from Inference}
\label{ssec:cause_decoding}

Decoding plays an important role in manifesting the capabilities of LLMs after pretraining and alignment. However, certain shortcomings in decoding strategies can lead to LLM hallucinations. 

\subsubsection{Imperfect Decoding Strategies}
\label{sssec:cause_decoding_strategies}

LLMs have demonstrated a remarkable aptitude for generating highly creative and diverse content, a proficiency that is critically dependent on the pivotal role of \textit{randomness} in their decoding strategies. Stochastic sampling \citep{fan2018hierarchical,  holtzman2019curious} is currently the prevailing decoding strategy employed by these LLMs.  The rationale for incorporating randomness into decoding strategies stems from the realization that high likelihood sequences often result in surprisingly low-quality text, which is called \textit{likelihood trap} \citep{stahlberg2019nmt, holtzman2019curious, meister2020if, zhang2020trading}.  The diversity introduced by the randomness in decoding strategies comes at a cost, as it is positively correlated with an increased risk of hallucinations \citep{dziri2021neural, chuang2023dola}. An elevation in the sampling temperature results in a more uniform token probability distribution, increasing the likelihood of sampling tokens with lower frequencies from the tail of the distribution. Consequently, this heightened tendency to sample infrequently occurring tokens exacerbates the risk of hallucinations \citep{aksitov2023characterizing}.

\subsubsection{Over-confidence}
\label{sssec:cause_over_confidence}
Prior studies in conditional text generation \citep{miao2021prevent, chen2022towards} have highlighted the issue of \textit{over-confidence} which stems from an excessive focus on the partially generated content, often prioritizing fluency at the expense of faithfully adhering to the source context.
While LLMs, primarily adopting the causal language model architecture, have gained widespread usage, the \textit{over-confidence} phenomenon continues to persist. During the generation process, the prediction of the next word is conditioned on both the language model context and the partially generated text. However, as demonstrated in prior studies \citep{voita2019analyzing, beltagy2020longformer, DBLP:journals/corr/abs-2307-03172}, language models often exhibit a localized focus within their attention mechanisms, giving priority to nearby words and resulting in a notable deficit in context attention \citep{DBLP:journals/corr/abs-2305-14739}.
Furthermore, this concern is further amplified in LLMs that exhibit a proclivity for generating lengthy and comprehensive responses. In such cases, there is even a heightened susceptibility to the risk of instruction forgetting \citep{chen2023improving, liu2023instruction}. This insufficient attention can directly contribute to faithfulness hallucinations, wherein the model outputs content that deviates from the original context.

\subsubsection{Softmax Bottleneck}
The majority of language models utilize a softmax layer that operates on the final layer's representation within the language model, in conjunction with a word embedding, to compute the ultimate probability associated with word prediction. Nevertheless, the efficacy of Softmax-based language models is impeded by a recognized limitation known as the \textit{Softmax bottleneck} \citep{yang2017breaking}, wherein the employment of softmax in tandem with distributed word embeddings constrains the expressivity of the output probability distributions given the context which prevents LMs from outputting the desired distribution.
Additionally, \citet{chang2022softmax} discovered that when the desired distribution within the output word embedding space exhibits multiple modes, language models face challenges in accurately prioritizing words from all the modes as the top next words, which also introduces the risk of hallucination.

\subsubsection{Reasoning Failure}
\label{sssec:cause_reasoning_failure}

Beyond the challenges with long-tail knowledge, effective utilization of knowledge is inextricably linked with reasoning capabilities. For instance, in multi-hop question-answering scenarios, even if the LLM possesses the necessary knowledge, it may struggle to produce accurate results if multiple associations exist between questions, due to its limitations in reasoning \citep{zheng2023does}. Furthermore, \citet{berglund2023reversal} unveiled a specific reasoning failure in LLMs termed the \textit{Reversal Curse}. Specifically, while the model can correctly answer when the question is formulated as "A is B", it exhibits a failed logical deduction when asked the converse "B is A". This discrepancy in reasoning extends beyond simple deductions. 

%% file: tables/bias_example.tex
\begin{table*}[!t]
\centering
\caption{Examples of Hallucinations from Misinformation and Biases. The table categorizes hallucinations arising from flawed data sources into imitative falsehoods, duplication bias, and social biases. Each category is accompanied by a premise outlining the data issue, user input, and the LLM’s hallucinatory output, and an explanation for the occurrence, aiding comprehension of these complex phenomena.}
\footnotesize
\renewcommand{\arraystretch}{1.2}
\begin{tabular}{m{1.4cm}m{2.5cm}m{3.0cm}m{2.6cm}m{2.7cm}}
\toprule
\textbf{Type}         & \textbf{Premise}    & \textbf{Input}       & \textbf{Model Ouput}                                                         & \textbf{Explanation}                                                 \\ 
         \midrule
\term{Imitative Falsehood}     & Within the LLM's training data, a frequently referenced but incorrect fact appears multiple times: \textit{Thomas Edison invented the light bulb.} &  Who is credited with the invention of the light bulb? & \textcolor{red}{Thomas Edison} is credited with the invention of the light bulb. & \textbf{While Edison improved the light bulb's design, he was not its sole inventor.} The model's answer reflects a misconception prevalent in its training data, exemplifying an imitative falsehood.    \\
\midrule
\term{Societal Biases}     &  In the model's training data, certain names may frequently be associated with specific nationalities.  & Provide a summary of the following news: Dr. Kim recently published a breakthrough paper on quantum computing. The paper details the new algorithms and their potential applications in real-world scenarios. It has been hailed as revolutionary by experts.   & Dr. Kim \textcolor{red}{from South Korea} recently published a paper on quantum computing, detailing new algorithms and their applications. & \textbf{The model erroneously added "South Korea" based on the name "Kim"} revealing its bias in linking certain names to specific nationalities, which can cause hallucinations in the summary. \\
\bottomrule
\end{tabular}
\label{data-induced_hallucination}
\end{table*}

%% file: tables/knowledge_boundary.tex
\begin{table*}[!t]
\centering
\caption{Example of Knowledge Boundary. Content marked in \textcolor{red}{Red} represents the hallucinatory output.}
\footnotesize
\renewcommand{\arraystretch}{1.2}
\begin{tabular}{m{2.5cm}m{2.7cm}m{3.5cm}m{3.8cm}}
\toprule
\textbf{Type} & \textbf{Input} & \textbf{Model Ouput} & \textbf{Explanation} \\ 
\midrule
\term{Long-tail Knowledge} & What causes phenylketonuria? & Phenylketonuria is caused by \textcolor{red}{a mutation in the hemoglobin gene, which results in an unusual interaction with hemoglobin}, affecting its ability to transport oxygen in the bloodstream. & The correct cause of phenylketonuria is mutations in the PAH gene. \textbf{Such specialized medical information is less frequently encountered in the training corpus, making LLMs struggle to memorize.} \\
\midrule
\term{Up-to-date Knowledge} & Who will be the Democratic Party candidate for the 2024 US presidential election? & \textcolor{red}{Michelle Obama has announced her candidacy and is leading in the polls, with significant endorsements from key Democratic figures such as Barack Obama and Bernie Sanders.} She has also outlined a policy platform focusing on healthcare reform and climate change initiatives. & After President Joe Biden announced his withdrawal from the race on July 21, 2024, he endorsed Kamala Harris as the Democratic Party candidate. \textbf{Such hallucinations arise because LLMs' parametric knowledge cannot be updated in real-time, leading to potential fabrication regarding current events.} \\
\midrule
\term{Copyright-sensitive Knowledge} & Describe the ending of the novel "The Secrets of the Alchemist" that was published last year.  & In the dramatic conclusion of "The Secrets of the Alchemist", \textcolor{red}{the protagonist discovers the true secret behind the philosopher's stone, leading to a peaceful resolution with all characters sharing in the wisdom.} & \textbf{The novel "The Secrets of the Alchemist" is under copyright protection, and LLMs have not been trained directly on such copyrighted materials.} Thus, the model’s output fabricates details about the book’s ending. \\
\bottomrule
\end{tabular}
\label{tab:knowledge_boundary}
\end{table*}

%% file: 4_Benchmark_and_Detection.tex
\section{Hallucination Detection and Benchmarks}
\label{sec:detection_and_benchmark}
The issue of hallucinations within LLMs has garnered considerable attention, raising concerns about the reliability of LLMs and their deployment in practical applications. 
As LLMs become increasingly adept at generating human-like text, accurately distinguishing between hallucinated versus factual content becomes increasingly vital.
Moreover, effectively measuring the level of hallucination in LLM is crucial for improving their reliability.
Thus, in this section, we delve into hallucination detection approaches (\S\ref{ssec: detection}) and benchmarks for assessing LLM hallucinations (\S\ref{ssec: benchmarks}).

\subsection{Hallucination Detection}
\label{ssec: detection}

Existing strategies for detecting hallucinations in LLMs can be categorized based on the type of hallucination: (1) factuality hallucination detection, which aims to identify factual inaccuracies in the model's outputs, and (2) faithfulness hallucination detection, which focuses on evaluating the faithfulness of model's outputs to the contextual information provided.

\subsubsection{Factuality Hallucination Detection}
\label{sssec: factuality_hallucination_detection}
Factuality hallucination detection involves assessing whether the output of LLMs aligns with real-world facts. Typical methods generally fall into two categories: \textit{fact-checking}, which involves verifying the factuality of the generated response against trusted knowledge sources, and \textit{uncertainty estimation}, which focuses on detecting factual inconsistency via internal uncertainty signals.

\paratitle{Fact-checking.}
Given that the output of LLMs is typically comprehensive and consists of multiple factual statements, the fact-checking approach is generally divided into two primary steps: (1) fact extraction, which involves extracting independent factual statements within the model's outputs (2) fact verification, which aims at verifying the correctness of these factual statements against trusted knowledge sources.
Depending on the type of knowledge sources employed for verification, fact-checking methodologies can be broadly categorized into two distinct parts: \textit{external retrieval} and \textit{internal checking}.

\begin{itemize}
	\item \textit{External retrieval}: The most intuitive strategy for fact verification is external retrieval. \citet{DBLP:journals/corr/abs-2305-14251} developed \textsc{FACTSCORE}, a fine-grained factual metric tailored for evaluating long-form text generation. It first decomposes the generation content into atomic facts and subsequently computes the percentage supported by reliable knowledge sources. Expanding on this concept, \citet{chern2023factool} proposed a unified framework that equips LLMs with the capability to identify factual inaccuracies by utilizing a collection of external tools dedicated to evidence gathering. In addition to retrieving supporting evidence solely based on decomposited claims, \citet{DBLP:journals/corr/abs-2306-13781} improved the retrieval process through query expansion. By combining the original question with the LLM-generated answer, they effectively addressed the issue of topic drift, ensuring that the retrieved evidence aligns with both the question and the LLM's response.
	\item \textit{Internal checking}: Given the extensive factual knowledge encoded in their parameters, LLMs have been explored as factual knowledge sources for fact-checking. \citet{dhuliawala2023chain} introduced the Chain-of-Verification (CoVe), where an LLM first generates verification questions for a draft response and subsequently leverages its parametric knowledge to assess the consistency of the answer against the original response, thereby detecting potential inconsistencies.\citet{kadavath2022language} and \citet{zhang2024self} calculates the probability $p(True)$ to assess the factuality of the response to a boolean question, relying exclusively on the model'sinternal knowledge. Additionally, \citet{li2024the} observed that most atomic statements are interrelated, some may serve as contextual backgrounds for others, which potentially leads to incorrect judgments. Thus, they instruct the LLM to directly predict hallucination judgments considering all factual statements. However, as LLMs are not inherently reliable factual databases \citep{zheng2024largelanguagemodelsreliable}, solely relying on LLMs' parametric knowledge for fact-checking may result in inaccurate assessments.
\end{itemize}

\paratitle{Uncertainty Estimation.}
While many approaches to hallucination detection rely on external knowledge sources for fact-checking, several methods have been devised to address this issue in zero-resource settings, thus eliminating the need for retrieval. The foundational premise behind these strategies is that the origin of LLM hallucinations is inherently tied to the model’s uncertainty.
Therefore, by estimating the uncertainty of the factual content generated by the model, it becomes feasible to detect hallucinations.  
The methodologies in uncertainty estimation can broadly be categorized into two approaches: based on \textit{LLM internal states} and \textit{LLM behavior}, as shown in Fig.~\ref{fig:factuality-detection-uncertainty}.

\input{figures/factuality_hallucination_detection_uncertainty}
\begin{itemize}
	\item \textit{LLM internal states}: The internal states of LLMs can serve as informative indicators of their uncertainty, often manifested through metrics like token probability or entropy. \citet{DBLP:journals/corr/abs-2307-03987} determined the model's uncertainty towards key concepts quantified by considering the minimal token probability within those concepts. The underlying rationale is that a low probability serves as a strong indicator of the model's uncertainty, with less influence from higher probability tokens present in the concept. Similarly, \citet{luo2023zero} employed a self-evaluation-based approach for uncertainty estimation by grounding in the rationale that a language model's ability to adeptly reconstruct an original concept from its generated explanation is indicative of its proficiency with that concept. By initially prompting the model to generate an explanation for a given concept and then employing constrained decoding to have the model recreate the original concept based on its generated explanation, the probability score from the response sequence can serve as a familiarity score for the concept. Furthermore, \citet{yao2023llm} interpreted hallucination through the lens of adversarial attacks. Utilizing gradient-based token replacement, they devised prompts to induce hallucinations. Notably, they observed that the first token generated from a raw prompt typically exhibits low entropy, compared to those from adversarial attacks. Based on this observation, they proposed setting an entropy threshold to define such hallucination attacks.
	\item \textit{LLM behavior}: However, when systems are only accessible via API calls \citep{2022chatgpt, 2023bard, 2023bing}, access to the output's token-level probability distribution might be unavailable. Given this constraint, several studies have shifted their focus to probing a model's uncertainty, either through natural language prompts \citep{DBLP:journals/corr/abs-2306-13063, kadavath2022language} or by examining its behavioral manifestations. For instance, by sampling multiple responses from an LLM for the same prompt, \citet{DBLP:journals/corr/abs-2303-08896} detected hallucinations via evaluating the consistency among the factual statements. However, these methods predominantly rely on direct queries that explicitly solicit information or verification from the model. \citet{DBLP:journals/corr/abs-2305-18248}, inspired by investigative interviews, advocated for the use of indirect queries. Unlike direct ones, these indirect counterparts often pose open-ended questions to elicit specific information. By employing these indirect queries, consistency across multiple model generations can be better evaluated. Beyond assessing uncertainty from the self-consistency of a single LLM's multiple generations, one can embrace a multi-agent perspective by incorporating additional LLMs. Drawing inspiration from legal cross-examination practices, \citet{DBLP:journals/corr/abs-2305-13281} introduced the LMvLM approach. This strategy leverages an examiner LM to question an examinee LM, aiming to unveil inconsistencies of claims during multi-turn interaction.
\end{itemize}

\subsubsection{Faithfulness Hallucination Detection}
\label{sssec: faithfulness_hallucination_detection}
Ensuring the faithfulness of LLMs to provide context or user instructions is pivotal for their practical utility in IR applications, from conversational search to interactive dialogue systems. 
We categorize existing hallucination detection metrics tailored to faithfulness into the following groups, with an overview shown in Fig.~\ref{fig:faithfulness-detection}: (1) Fact-based (2) Classifier-based (3) QA-based (4) Uncertainty-based (5) LLM-based.
\input{figures/faithful_detection}

\paratitle{Fact-based Metrics.} In the realm of assessing faithfulness, one of the most intuitive methods involves measuring the overlap of pivotal facts between the generated content and the source content. Given the diverse manifestations of facts, faithfulness can be measured based on \textit{n-gram}, \textit{entities}, and \textit{relation triples}.
Traditional \textit{n-gram-based} metrics, such as BLEU \citep{papineni2002bleu}, ROUGE \citep{lin2004rouge} and PARENT-T \citep{wang2020towards}, typically fall short in differentiating the nuanced discrepancies between the generated content and the source content \citep{maynez2020faithfulness}. 
\textit{Entity-based} metrics \citep{nan2021entity} make a step further by calculating the overlap of entities, as any omission or inaccurate generation of these key entities could lead to an unfaithful response. 
Notably, even if entities match, the relations between them might be erroneous. Thus, \textit{relation-based} metrics \citep{goodrich2019assessing} focus on the overlap of relation tuples and introduce a metric that computes the overlap of relation tuples extracted using trained end-to-end fact extraction models.

\paratitle{Classifier-based Metrics.} Beyond computing fact overlap, another straightforward approach to assessing the faithfulness of the model generation involves utilizing classifiers trained on data from related tasks such as natural language inference (NLI) and fact-checking, or data comprised of synthetically task-specific hallucinated and faithful content.
A foundational principle for assessing the faithfulness of generated text is anchored on the idea that genuinely faithful content should inherently be entailed by its source content. 
In line with this, numerous studies \citep{falke2019ranking, maynez2020faithfulness} have trained classifiers on NLI datasets to identify factual inaccuracies, especially in the context of abstract summarization. However, \citet{mishra2021looking} highlighted that the mismatch in input granularity between conventional NLI datasets and inconsistency detection datasets limits their applicability for effectively detecting inconsistencies. Building on this, more advanced studies have proposed methods such as fine-tuning on adversarial datasets \citep{barrantes2020adversarial}, decomposing the entailment decisions at the dependency arc level \citep{goyal2020evaluating}, and segmenting documents into sentence units then aggregating scores between sentence pairs \citep{laban2022summac}. 
While using data from related tasks to fine-tune the classifier has shown promise in evaluating faithfulness, it's essential to recognize the inherent gap between related tasks and the downstream task. The scarcity of annotated data further constrains their applicability. In response to this challenge, a surge of research explores leveraging data-augmentation methods to construct synthetical data for fine-tuning the classifier, either by rule-based perturbation \citep{dziri2021evaluating, kryscinski2019evaluating, santhanam2021rome} or generation \citep{zhou2020detecting}.

\paratitle{QA-based Metrics.}
In contrast to classifier-based metrics, QA-based metrics \citep{durmus2020feqa, wang2020asking, scialom2021questeval, honovich2021q} have recently garnered attention for their enhanced ability to capture information overlap between the model's generation and its source. These metrics operate by initially selecting target answers from the information units within the LLM's output, and then questions are generated by the question-generation module. The questions are subsequently used to generate source answers based on the user context. Finally, the faithfulness of the LLM's responses is calculated by comparing the matching scores between the source and target answers. 
Although these methodologies share a common thematic approach, they exhibit variability in aspects like answer selection, question generation, and answer overlap, leading to diverse performance outcomes. Building on this foundational work, \citet{fabbri2021qafacteval} conducted an in-depth evaluation of the components within QA-based metrics, yielding further enhancements in faithfulness evaluation. 

\paratitle{Uncertainty-based Metrics.}
 Drawing parallels with the uncertainty-based approaches employed for detecting factuality hallucinations (\S\ref{sssec: factuality_hallucination_detection}), the application of uncertainty estimation in assessing faithfulness has been widely explored, typically characterized by entropy and log-probability.
For entropy-based uncertainty, \citet{xiao2021hallucination} has revealed a positive correlation between hallucination likelihood in data-to-text generation and predictive uncertainty, which is estimated by deep ensembles \citep{lakshminarayanan2017simple}. In a related vein, \citet{guerreiro2022looking} leveraged the variance in hypotheses yielded by Monte Carlo Dropout \citep{gal2016dropout} as an uncertainty measure within neural machine translation. More recently, \citet{van2022mutual} employed conditional entropy \citep{xu2020understanding} to assess model uncertainty in abstractive summarization.
Regarding log-probability, it can be applied at different levels of granularity, such as word or sentence level.
Notably, several studies \citep{guerreiro2022looking, yuan2021bartscore, fu2023gptscore} have adopted length-normalized sequence log-probability to measure model confidence.
Furthermore, considering the hallucinated token can be assigned high probability when the preceding context contains the same hallucinated information, \citet{zhang2023enhancing} focused on the most informative and important keywords and introduced a penalty mechanism to counteract the propagation of hallucinated content.

\paratitle{LLM-based Judgement.}
Recently, the remarkable instruction-following ability of LLMs has underscored their potential for automatic evaluation \citep{chiang2023can, liu2023gpteval, wang2023chatgpt}. Exploiting this capability, researchers have ventured into novel paradigms for assessing the faithfulness of model-generated content \citep{luo2023chatgpt, laban2023llms, adlakha2023evaluating, gao2023human, jain2023multi}. By providing LLMs with concrete evaluation guidelines and feeding them both the model-generated and source content, they can effectively assess faithfulness. The final evaluation output can either be a binary judgment on faithfulness \citep{luo2023chatgpt} or a k-point Likert scale indicating the degree of faithfulness \citep{gao2023human}. For prompt selection, evaluation prompt can either be direct prompting, chain-of-thought prompting \citep{adlakha2023evaluating}, using in-context-learning \citep{jain2023multi} or allowing the model to generate evaluation results accompanying with explanations \citep{laban2023llms}.

\subsection{Hallucination Benchmarks}
\label{ssec: benchmarks}

In this section, we present a comprehensive overview of existing hallucination benchmarks, which can be categorized into two primary domains: Hallucination Evaluation Benchmarks (\S\ref{sssec:evaluation_benchmark}), which assess the extent of hallucinations generated by existing cutting-edge LLMs, and Hallucination Detection Benchmarks (\S\ref{sssec:detection_benchmark}), designed specifically to evaluate the performance of existing hallucination detection methods. Collectively, these benchmarks establish a unified framework, enabling a nuanced and thorough exploration of hallucinatory patterns in LLMs.

\input{tables/benchmark}
\subsubsection{Hallucination Evaluation Benchmarks}
\label{sssec:evaluation_benchmark} 
Hallucination evaluation benchmarks are devised to quantify the tendency of LLMs to generate hallucinations, particularly emphasizing factual inaccuracies and inconsistency from the given contexts. 
Given the adeptness of LLMs at memorizing high-frequency count knowledge, the primary focus of current hallucination evaluation benchmarks targets long-tailed knowledge and challenging questions that can easily elicit imitative falsehood.
As for evaluating, these benchmarks typically utilize multiple choice QA, where performance is measured through accuracy metrics, or generative QA, evaluated either through human judgment or scores given by proxy models.

\paratitle{Long-tail Factual Knowledge.}
The selection criteria for gathering long-tail factual question-answering samples typically include the frequency of appearance, recency, and specific domains.
Regarding the frequency of appearance, benchmarks such as PopQA \citep{DBLP:conf/acl/MallenAZDKH23} and Head-to-Tail \citep{sun2023head} are constructed based on entity popularity derived directly from Wikipedia.
Considering that world knowledge is constantly evolving, it becomes crucial to validate the LLM's factuality concerning the current world. Among benchmarks characterized by ever-changing, REALTIMEQA \citep{kasai2022realtime} and FreshQA \citep{vu2023freshllms} stands out. 
REALTIMEQA offers real-time, open-domain multiple-choice questions that are regularly updated to reflect the latest developments. These questions are derived from newly published news articles, encompassing a broad spectrum of topics, including politics, business, sports, and entertainment.
Similarly, FreshQA challenges LLMs with questions designed to represent varying degrees of temporal change—categorized into never-changing, slow-changing, and fast-changing world knowledge. This benchmark is further enriched by including questions based on false premises, requiring debunking, thus comprising a total of 600 meticulously hand-crafted questions. 
Moreover, long-tail knowledge often pertains to specific domains. 
For instance, Med-HALT \citep{umapathi2023med} is distinguished by its focus on the medical domain, challenging LLMs with multiple-choice questions derived from a variety of countries.
Additionally, \citet{malaviya2023expertqa} collected expert-curated questions across 32 fields of study, resulting in a high-quality long-form QA dataset with 2,177 questions.

\paratitle{Imitative Falsehood Knowledge.}
Imitative falsehood knowledge is specifically designed to challenge LLMs through adversarial prompting. This approach crafts questions in such a way that they are prone to misleading LLMs due to false beliefs or misconceptions.
The two most representative benchmarks are TruthfulQA \citep{DBLP:conf/acl/LinHE22} and HalluQA \citep{cheng2023evaluating}.
TruthfulQA comprises 817 questions that span 38 diverse categories, such as health, law, finance, and politics. Crafted using an adversarial methodology, it aims to elicit "imitative falsehoods"—misleading responses that models might generate due to their frequent presence in training data. The benchmark is divided into two parts, one of which contains manually curated questions that were further refined by filtering out those correctly answered by GPT-3, resulting in 437 filtered questions. The other part includes 380 unfiltered non-adversarial questions. 
Drawing from the construction approach of TruthfulQA, HalluQA is crafted to specifically assess hallucinations in Chinese LLMs, focusing on imitative falsehoods and factual errors. The benchmark comprises 450 handcrafted adversarial questions across 30 domains and is categorized into two parts. The misleading section captures questions that successfully deceive GLM-130B, while the knowledge section retains questions that both ChatGPT and Puyu consistently answer incorrectly.
To comprehensively evaluate LLM hallucinations across various domains, \citet{li2024the} constructed an upgraded hallucination evaluation benchmark, HaluEval 2.0, based on \citep{li2023halueval}. This benchmark includes 8,770 questions that LLMs are prone to hallucination, across five domains: biomedicine, finance, science, education, and open domain.

\subsubsection{Hallucination Detection Benchmarks}
\label{sssec:detection_benchmark}
For hallucination detection benchmarks, most prior studies have primarily concentrated on task-specific hallucinations, such as abstractive summarization \citep{kryscinski2019evaluating, wang2020asking, maynez2020faithfulness, fabbri2021summeval, goyal2021annotating, pagnoni2021understanding}, data-to-text\citep{tian2019sticking, parikh2020totto}, and machine translation \citep{zhou2020detecting}. However, the content generated in these studies often originates from models with lesser capabilities, such as BART \citep{lewis2019bart} and PEGASUS \citep{zhang2020pegasus}. As a result, they may not accurately reflect the effectiveness of hallucination detection strategies, underlining the necessity for a significant shift toward developing benchmarks that encapsulate more complex scenarios reflective of the era of LLMs.

For example, SelfCheckGPT-Wikibio \citep{miao2023selfcheck} offers a sentence-level dataset created by generating synthetic Wikipedia articles with GPT-3, manually annotated for factuality, highlighting the challenge of detecting hallucinations in the biography domain. Complementing this, HaluEval \citep{li2023halueval} combines automated generation with human annotation to evaluate LLMs' ability to recognize hallucinations across 5,000 general user queries and 30,000 task-specific samples, leveraging a "sampling-then-filtering" approach. 
Building upon existing research predominantly focused on short documents, BAMBOO \citep{dong2023bamboo} and ScreenEval \citep{lattimer2023fast} extend the scope in long-form hallucination detection.
Further, FELM \citep{chen2023felm}, distinguishes itself by assessing factuality across diverse domains including world knowledge, science, and mathematics, producing 817 samples annotated for various facets of factual accuracy, thereby addressing the need for cross-domain evaluation of factuality in LLM-generated content. On a different note, PHD \citep{yang2023new}, shifts the focus towards passage-level detection of non-factual content by analyzing entities from Wikipedia, thus offering a nuanced view on the knowledge depth of LLMs.
RealHall \citep{friel2023chainpoll} and SAC$^3$ \citep{zhang2023sac3} align closely with real-world applications focusing on open-domain question-answering, whereas LSum \citep{feng2023improving} concentrating on summarization tasks.

%% file: figures/factuality_hallucination_detection_uncertainty.tex
\begin{figure*}[t]
    \centering
    \includegraphics[width=1.0\textwidth]{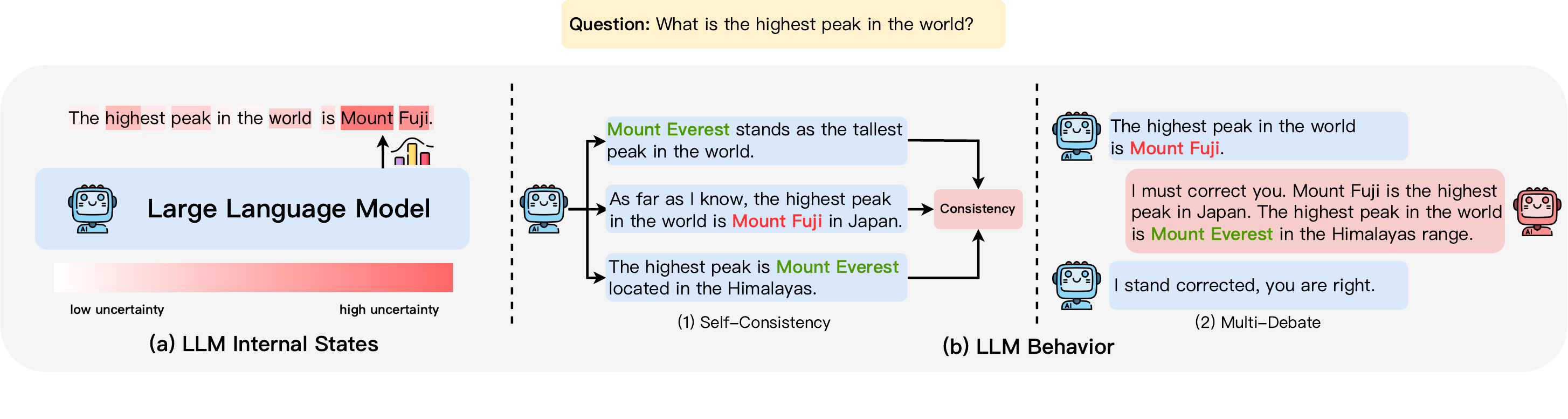}
    \caption{Taxonomy of Uncertainty Estimation Methods in Factual Hallucination Detection, featuring \textbf{a) LLM Internal States} and \textbf{b) LLM Behavior}, with LLM Behavior encompassing two main categories: Self-Consistency and Multi-Debate.}
    \label{fig:factuality-detection-uncertainty}
\end{figure*}

%% file: figures/faithful_detection.tex
\begin{figure*}[!t]
    \centering
      \includegraphics[width=0.9\textwidth]{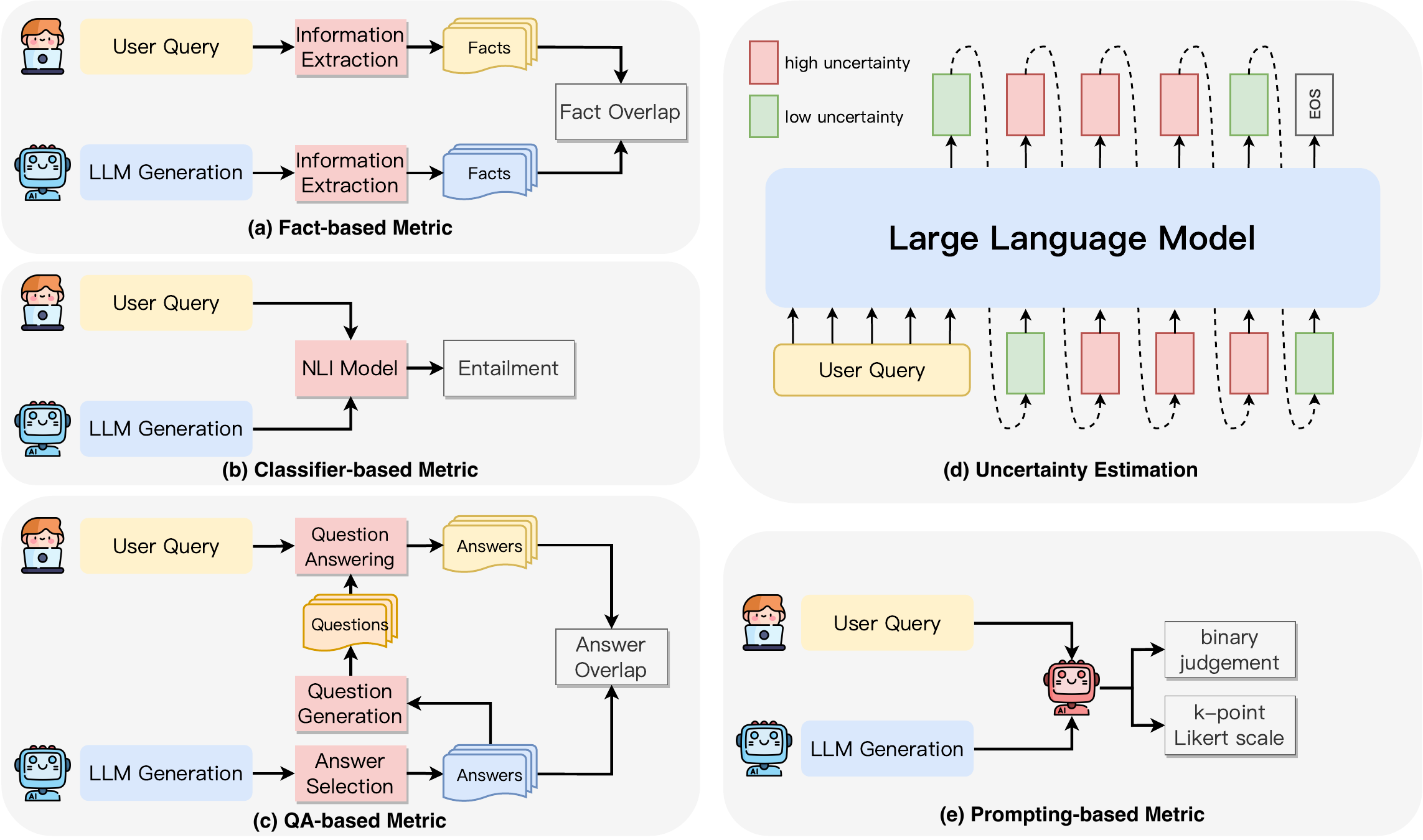}
    \caption{The illustration of detection methods for faithfulness hallucinations: \textbf{a) Fact-based Metrics}, which assesses faithfulness by measuring the overlap of facts between the generated content and the source content; \textbf{b) Classifier-based Metrics}, utilizing trained classifiers to distinguish the level of entailment between the generated content and the source content; \textbf{c) QA-based Metrics}, employing question-answering systems to validate the consistency of information between the source content and the generated content; \textbf{d) Uncertainty Estimation}, which assesses faithfulness by measuring the model’s confidence in its generated outputs; \textbf{e) Prompting-based Metrics}, wherein LLMs are induced to serve as evaluators, assessing the faithfulness of generated content through specific prompting strategies.}
	\label{fig:faithfulness-detection}
\end{figure*}

%% file: tables/benchmark.tex
\begin{table*}[!ht]
\centering
\caption{An overview of existing hallucination benchmarks. For Attribute, \textbf{Factuality} and \textbf{Faithfulness} represent whether the benchmark is used to evaluate LLM's factuality or to detect faithfulness hallucination, and \textbf{Manual} represents whether the inputs in the data are handwritten.}
\scriptsize
\renewcommand\arraystretch{1.2}
\resizebox{\textwidth}{!}{
\begin{tabular}{clllccccccc}
\toprule
                                      &                                     &                                     &                                    & \multicolumn{3}{c}{\textbf{Attribute}}                                                 & \multicolumn{4}{c}{\textbf{Task}}                                                                                                                           \\
                                      \cmidrule(lr){5-7} \cmidrule(lr){8-11}
\multirow{-2}{*}{\textbf{Benchmark}}       & \multirow{-2}{*}{\textbf{Datasets}}     & \multirow{-2}{*}{\textbf{Data Size}}      & \multirow{-2}{*}{\textbf{Language}} & \textbf{Factuality}             & \textbf{Faithfulness}                             & \textbf{Manual}          & \textbf{Task Type}                 & \textbf{Input}                             & \textbf{Label}       & \textbf{Metric} \\
\midrule
                                      &                                     &                          &                                    &                            &                                             &                        & Generative QA                               &                                             &                     &                                                                        \\
\multirow{-2}{*}{\makecell{TruthfulQA \\ \cite{DBLP:conf/acl/LinHE22}}}                                 & \multirow{-2}{*}{-}                                  & \multirow{-2}{*}{817}                                & \multirow{-2}{*}{English}                               & \multirow{-2}{*}{\greencheck}                        & \multirow{-2}{*}{\redcross}                   & \multirow{-2}{*}{\greencheck}                      & Multi-Choice QA                             & \multirow{-2}{*}{Question}                   & \multirow{-2}{*}{Answer}                   & \multirow{-2}{*}{\makecell{ LLM-Judge \& \\ Human}}                                                                   \\
\midrule
                                      &                                     &                          &                                    &                            &                                             &                        & Multi-Choice QA                               &                                             &                     &  Acc                                                                     \\
\multirow{-2}{*}{\makecell{REALTIMEQA \\ \cite{kasai2022realtime}}}               & \multirow{-2}{*}{-}     & \multirow{-2}{*}{Dynamic}             & \multirow{-2}{*}{English}        & \multirow{-2}{*}{\greencheck} & \multirow{-2}{*}{\redcross} & \multirow{-2}{*}{\greencheck}    & Generative QA & \multirow{-2}{*}{Question} & \multirow{-2}{*}{Answer} & EM \& F1        \\
\midrule
\makecell{SelfCheckGPT-Wikibio \\ \cite{miao2023selfcheck}}                & -                & 1,908  & English             & \redcross      & \greencheck & \redcross    & Detection          & \makecell{Paragraph \& \\ Concept} & Passage & AUROC                              \\
\midrule
                         & Task-specific                                 & 30,000                                & English                               & \redcross                        & \greencheck                   & \redcross                      & Detection                            & Query                   & Response                   & Acc \\
                         
\multirow{-2}{*}{\makecell{HaluEval \\ \cite{li2023halueval}}}                               & General                                 & 5,000                                & English                               & \redcross                        & \greencheck                   & \redcross                      & Detection                            & Task Input                   & Response                   & Acc                                                             \\
\midrule
\makecell{Med-HALT \\ \cite{umapathi2023med}}                                  & -                                 & 4,916                                & Multilingual                               & \greencheck                       & \redcross                   & \redcross                      & Multi-Choice QA                            & Question                   &  Choice                  & \makecell{Pointwise Score \\ \& Acc}                                                          \\
\midrule
                 & Wiki-FACTOR      & 2,994                & English             & \greencheck      & \redcross & \redcross    & Multi-Choice QA & Question & Answer & likelihood                             \\
                                      
\multirow{-2}{*}{\makecell{FACTOR \\ \cite{muhlgay2023generating}}}                  & News-FACTOR      & 1,036                & English             & \greencheck      & \redcross & \redcross    & Multi-Choice QA & Question & Answer & likelihood                             \\
\midrule
        & SenHallu                & 200  & English             & \redcross      & \greencheck & \redcross    & Detection          & Paper & Summary & \makecell{P \& R \& F1}                             \\
                                      
\multirow{-2}{*}{\makecell{BAMBOO \\ \cite{dong2023bamboo}}}                & AbsHallu                & 200  & English             & \redcross      & \greencheck & \redcross    & Detection          & Paper & Summary & \makecell{P \& R \& F1}                              \\
\midrule
\makecell{ChineseFactEval \\ \cite{wang2023chinesefacteval}}                            & -                             & 125                                & Chinese                               & \greencheck                        & \redcross                   & \greencheck                      & Generative QA                            & Question                   & -                   & Score                                                                    \\
\midrule
                           & Misleading                                & 175                               & Chinese                               & \greencheck                        & \redcross                                           & \greencheck                   & Generative QA                            & Question                   & Answer                   & LLM-Judge                                                                   \\
                           
                           & Misleading-hard                                 & 69                               & Chinese                               & \greencheck                        & \redcross                                           & \greencheck                   & Generative QA                            & Question                   & Answer                   & LLM-Judge                                                                   \\
                           
\multirow{-3}{*}{\makecell{HaluQA \\ \cite{cheng2023evaluating}}}                            & Knowledge                                 & 206                               & Chinese                               & \greencheck                        & \redcross                                           & \greencheck                   & Generative QA                            & Question                   & Answer                   & LLM-Judge                                                                   \\
\midrule
                              & Never-changing                                 & 150                                & English                               & \greencheck                        & \redcross                   & \greencheck                      & Generative QA                        & Question                   & Answer                   & Human                                                                   \\
                              
                               & Slow-changing                                 & 150                                & English                               & \greencheck                        & \redcross                   & \greencheck                      & Generative QA                        & Question                   & Answer                   & Human                                                                   \\
                               
                               & Fast-changing                                 & 150                                & English                               & \greencheck                        & \redcross                   & \greencheck                      & Generative QA                        & Question                   & Answer                   & Human                                                                   \\
                               
\multirow{-4}{*}{\makecell{FreshQA \\ \cite{vu2023freshllms}}}                               & False-premise                                 & 150                                & English                               & \greencheck                        & \redcross                   & \greencheck                      & Generative QA                        & Question                   & Answer                   & Human                                                                   \\
\midrule
\makecell{FELM \\ \cite{chen2023felm}}                & -                & 3,948  & English             & \greencheck      & \greencheck & \redcross    & Detection          & Question & \makecell{Response} & \makecell{Balanced \\ Acc \& F1}                                \\
\midrule
               & PHD-LOW                & 100  & English             & \redcross      & \greencheck & \redcross    & Detection          & Entity & \makecell{Response} & \makecell{P \& R \& F1} \\
                                      
             & PHD-Meidum                & 100  & English             & \redcross      & \greencheck & \redcross    & Detection          & Entity & \makecell{Response} & \makecell{P \& R \& F1}                   \\
                                      
\multirow{-3}{*}{\makecell{PHD \\ \cite{yang2023new}}}                & PHD-High                & 100  & English             & \redcross      & \greencheck & \redcross    & Detection          & Entity & \makecell{Response} & \makecell{P \& R \& F1}                                \\
\midrule
\makecell{ScreenEval \\ \cite{lattimer2023fast}}                & -                & 52  & English             & \redcross      & \greencheck & \redcross    & Detection          & Document  & Summary & AUROC                                \\
\midrule
                & COVID-QA                & N/A  & English             & \redcross      & \greencheck & \redcross    & Detection          & \makecell{Question} & Answer & AUROC                                \\
                 
                & DROP                & N/A  & English             & \redcross      & \greencheck & \redcross    & Detection          & \makecell{Question} & Answer & AUROC                                \\
                 
                & Open Assistant                & N/A  & English             & \redcross      & \greencheck & \redcross    & Detection          & Question & Answer & AUROC                                \\
                 
\multirow{-3}{*}{\makecell{RealHall \\ \cite{friel2023chainpoll}}}                & TriviaQA                & N/A  & English             & \redcross      & \greencheck & \redcross    & Detection          & Question & Answer & AUROC                                \\
\midrule
\makecell{LSum \\ \cite{feng2023improving}}                &  -               & 6,166  & English             & \redcross      & \greencheck & \redcross    & Detection          & Document & Summary & Balanced Acc                              \\
\midrule
                                      &      HotpotQA                               &    250                      &    English                                &    \redcross                        &   \greencheck                             &    \redcross                    &      Detection                           &   Question                                          &      Answer               &    AUROC                                                                   \\
                                      
\multirow{-2}{*}{\makecell{SAC$^3$ \\ \cite{zhang2023sac3}}}                & NQ-Open                & 250  & English             & \redcross      & \greencheck & \redcross    & Detection          & Question & Answer & AUROC                              \\
\midrule
                              & Biomedicine                               & 1,535                                & English                               & \greencheck                        & \redcross                   & \redcross                      & Generative QA                        & Question                   & Answer                   & MiHR \& MaHR                                                                   \\
							& Finance                                 & 1,125                                & English                               & \greencheck                        & \redcross                   & \redcross                      & Generative QA                        & Question                   & Answer                   & MiHR \& MaHR                                                                   \\
                               & Science                                 & 1,409                                & English                               & \greencheck                        & \redcross                   & \redcross                      & Generative QA                        & Question                   & Answer                   & MiHR \& MaHR                                                                   \\
                               
                               & Education                                 & 1,701                                & English                               & \greencheck                        & \redcross                   & \redcross                      & Generative QA                        & Question                   & Answer                   & MiHR \& MaHR                                                                   \\
                               
\multirow{-5}{*}{\makecell{HaluEval 2.0 \\ \cite{li2024the}}}                               & Open domain                                 & 3,000                                & English                               & \greencheck                        & \redcross                   & \redcross                      & Generative QA                        & Question                   & Answer                   & MiHR \& MaHR                                                                   \\
\bottomrule
\end{tabular}
} 
\end{table*}

%% file: 5_Mitigating.tex
\section{Hallucination Mitigation}
\label{sec:mitigating}

In this section, we present a comprehensive review of contemporary methods aimed at mitigating hallucinations in LLMs. Drawing from insights discussed in \textit{Hallucination Causes} (\S\ref{sec:causes}), we systematically categorize these methods based on the underlying causes of hallucinations. Specifically, we focus on approaches addressing \textit{Data-related Hallucinations} (\S\ref{ssec:mitigating_data}), \textit{Training-related Hallucinations} (\S\ref{ssec:mitigating_training}) and \textit{Inference-related Hallucinations} (\S\ref{ssec:mitigating_inference}), each offering tailored solutions to tackle specific challenges inherent to their respective cause.

\subsection{Mitigating Data-related Hallucinations}
\label{ssec:mitigating_data}

As analyzed in \S\ref{ssec:cause_data}, data-related hallucinations generally emerge as a byproduct of misinformation, biases, and knowledge gaps, which are fundamentally rooted in the pre-training data. Several methods are proposed to mitigate such hallucinations, primarily categorized into three distinct parts: (1) \textit{data filtering} aiming at selecting high-quality data to avoid introducing misinformation and biases, (2) \textit{model editing} focusing on injecting up-to-date knowledge by editing model's parameters, and (3) \textit{retrieval-augmented generation} leveraging external non-parametric database for knowledge supplying.

\subsubsection{Data Filtering}
\label{sssec:mitigating_data_filtering}
To reduce the presence of misinformation and biases, an intuitive approach involves the careful selection of high-quality pre-training data from reliable sources. In this way, we can ensure the factual correctness of data while also minimizing the introduction of social biases. 
As early as the advent of GPT-2, \citet{radford2019language} underscored the significance of exclusively scraping web pages that had undergone rigorous curation and filtration by human experts. However, as pre-training datasets continue to scale, manual curation becomes a challenge. Given that academic or specialized domain data is typically factually accurate, gathering high-quality data emerges as a primary strategy. Notable examples include \textit{the Pile} \citep{gao2020pile} and “textbook-like” data sources \citep{gunasekar2023textbooks, li2023textbooks}. Additionally, up-sampling factual data during the pre-training phase has been proven effective in enhancing the factual correctness of LLMs \citep{touvron2023llama}, thus alleviating hallucination.

In addition to strictly controlling the source of data, deduplication serves as a crucial procedure. Existing practices typically fall into two categories: exact duplicates and near-duplicates. For exact duplicates, the most straightforward method involves exact substring matching to identify identical strings. However, given the vastness of pre-training data, this process can be computationally intensive, a more efficient method utilizes the construction of a suffix array \citep{manber1993suffix}, enabling effective computation of numerous substring queries in linear time. Regarding near-duplicates, the identification often involves approximate full-text matching, typically utilizing hash-based techniques to identify document pairs with significant n-gram overlap. Furthermore, MinHash \citep{broder1997resemblance} stands out as a prevalent algorithm for large-scale deduplication tasks \citep{gyawali2020deduplication}. Additionally, SemDeDup \citep{abbas2023semdedup} makes use of embeddings from pre-trained models to identify semantic duplicates, which refers to data pairs with semantic similarities but not identical.

\paratitle{Discussion.}
Since data filtering works directly at the source of hallucinations, it effectively mitigates hallucinations by ensuring the use of high-quality, factually accurate sources.
Despite its effectiveness, the efficiency and scalability of current data filtering methods pose significant challenges as data volumes expand.
Additionally, these methods often overlook the influence of LLM-generated content, which can introduce new risks and inaccuracies.
To advance, future research must focus on developing more efficient, automated data filtering algorithms that can keep pace with the rapid expansion of datasets and the complexities of LLM-generated content.

\subsubsection{Model Editing}
\label{sssec:mitigating_knowledge_editing}
Model editing \citep{sinitsin2020editable, wang2023knowledge, zhang2024comprehensive} has garnered rising attention from researchers, which aims to rectify model behavior by incorporating additional knowledge. Current model editing techniques can be categorized into two classes: \textit{locate-then-edit} and \textit{meta-learning}.

\paratitle{Locate-then-edit.}
 Locate-then-edit methods \citep{dai2021knowledge, meng2022locating}  consist of two stages, which first locate the “buggy” part of the model parameters and then apply an update to them to alter the model's behavior. 
 For example, ROME \citep{meng2022locating} located the edits-related layer by destroying and subsequently restoring the activations and then updates the parameters of FFN in a direct manner to edit knowledge. MEMIT \citep{meng2022mass} employed the same knowledge locating methods as ROME, enabling the concurrent updating of multiple layers to facilitate the simultaneous integration of thousands of editing knowledge. 
 However, \citet{yao2023editing} found that these methods lack non-trivial generalization capabilities and varying performance and applicability to different model architectures. The best-performing methods ROME and MEMIT empirically only work well on decoder-only LLMs. 
 
 \paratitle{Meta-learning.}
 Meta-learning methods \citep{de2021editing, mitchell2021fast} train an external hyper-network to predict the weight update of the original model. Nevertheless, meta-learning methods often require additional training and memory cost, where MEND \citep{mitchell2021fast} utilized a low-rank decomposition with a specialized design to reduce the size of hyper-networks.
 Notably, MEND would exhibit a cancellation effect, where parameter shifts corresponding to different keys significantly counteract each other. MALMEN \citep{tan2024massive} further addressed this issue by framing the parameter shift aggregation as a least squares problem rather than a simple summation, thereby greatly enhancing its capacity for extensive editing.
 While these methods can fine-grainedly adjust the behavior of the model, modifications to the parameters could have a potentially harmful impact on the inherent knowledge of the model.

\paratitle{Discussion.}
Model editing provides a precise way to mitigate hallucinations induced by specific misinformation without extensive retraining. However, these methods struggle with large-scale updates and can adversely affect the model's overall performance, particularly when continuous edits are applied.
Consequently, future research should focus on improving model editing to handle large-scale knowledge updates more efficiently and address hallucinations caused by social biases.

\subsubsection{Retrieval-Augmented Generation}
\label{sssec:mitigating_rag}
Typically, retrieval-augmented generation (RAG) \citep{lewis2020retrieval, guu2020retrieval, shuster2021retrieval} follows a retrieve-then-read pipeline, where relevant knowledge is firstly retrieved by a \textit{retriever} \citep{DBLP:conf/emnlp/KarpukhinOMLWEC20} from external sources, and then the final response is generated by a \textit{generator} conditioning on both user query and retrieved documents. 
By decoupling external knowledge from LLM, RAG can effectively alleviate the hallucination caused by the knowledge gap without affecting the performance of LLM. Common practices can be divided into three parts, as shown in Fig~\ref{fig:retrieval}: \textit{one-time retrieval}, \textit{iterative retrieval}, and \textit{post-hoc retrieval}, depending on the timing of retrieval.
\input{figures/retrieval}

\paratitle{One-time Retrieval.}
One-time retrieval aims to directly prepend the external knowledge obtained from a single retrieval to the LLMs' prompt. \citet{DBLP:journals/corr/abs-2302-00083} introduced In-context RALM, which entails a straightforward yet effective strategy of prepending chosen documents to the input text of LLMs.  
Beyond conventional knowledge repositories such as Wikipedia, ongoing research endeavors have explored alternative avenues, specifically the utilization of knowledge graphs (KGs). These KGs serve as a pivotal tool for prompting LLMs, facilitating their interaction with the most recent knowledge, and eliciting robust reasoning pathways \citep{DBLP:journals/corr/abs-2308-09729, DBLP:journals/corr/abs-2308-10173, DBLP:journals/corr/abs-2306-04136}. \citet{DBLP:journals/corr/abs-2307-03987} introduce the Parametric Knowledge Guiding (PKG) framework, enhancing LLMs with domain-specific knowledge. PKG employs a trainable background knowledge module, aligning it with task knowledge and generating relevant contextual information. 
 
\paratitle{Iterative Retrieval.}
 When confronted with intricate challenges like multi-step reasoning \citep{DBLP:conf/emnlp/Yang0ZBCSM18} and long-form question answering \citep{DBLP:conf/acl/FanJPGWA19, DBLP:conf/emnlp/StelmakhLDC22}, traditional one-time retrieval may fall short. 
Addressing these demanding information needs, recent studies have proposed iterative retrieval, which allows for continuously gathering knowledge throughout the generation process. 
Recognizing the substantial advancements chain-of-thought prompting \citep{wei2022chain} has brought to LLMs in multi-step reasoning, numerous studies \cite{yao2022react, DBLP:conf/acl/TrivediBKS23, he2022rethinking} try to incorporate external knowledge at each reasoning step and further guide retrieval process based on ongoing reasoning, reducing factual errors in reasoning chains. Building upon chain-of-thought prompting, \citet{press2022measuring} introduced \textit{self-ask}. Diverging from the conventional continuous, undelineated chain-of-thought prompting, \textit{self-ask} delineates the question it intends to address at each step, subsequently incorporating a search action based on the follow-up question. Instead of solely depending on chain-of-thought prompting for retrieval guidance, both \citet{feng2023retrieval} and \citet{DBLP:journals/corr/abs-2305-15294} employed an iterative retrieval-generation collaborative framework, where a model's response serves as an insightful context to procure more relevant knowledge, subsequently refining the response in the succeeding iteration. 
 Beyond multi-step reasoning tasks, \citet{DBLP:journals/corr/abs-2305-06983} shifted their emphasis to long-form generation. They proposed an active retrieval augmented generation framework, which iteratively treats the upcoming prediction as a query to retrieve relevant documents. If the prediction contains tokens of low confidence, the sentence undergoes regeneration. In addition to using iterative retrieval to improve intermediate generations, \citet{DBLP:journals/corr/abs-2305-13669} presented MixAlign, which iteratively refines user questions using model-based guidance and seeking clarifications from users, ultimately enhancing the alignment between questions and knowledge.
 
\paratitle{Post-hoc Retrieval.}
Beyond the traditional \textit{retrieve-then-read} paradigm, a line of work has delved into post-hoc retrieval, refining LLM outputs through subsequent retrieval-based revisions. 
To enhance the trustworthiness and attribution of LLMs, \citet{DBLP:conf/acl/GaoDPCCFZLLJG23} adopted the \textit{research-then-revise} workflow, which initially research relevant evidence and subsequently revise the initial generation based on detected discrepancies with the evidence. Similarly, \citet{DBLP:conf/acl/ZhaoLJQB23} introduced the \textit{verify-and-edit} framework to enhance the factual accuracy of reasoning chains by incorporating external knowledge. For reasoning chains that show lower-than-average consistency, the framework generates verifying questions and then refines the rationales based on retrieved knowledge, ensuring a more factual response. \citet{yu2023improving} enhanced the post-hoc retrieval method through diverse answer generation.  Instead of generating just a single answer, they sample various potential answers, allowing for a more comprehensive retrieval feedback. Additionally, by employing an ensembling technique that considers the likelihood of the answer before and after retrieval, they further mitigate the risk of misleading retrieval feedback.
 
\paratitle{Discussion.}
One crucial advantage of retrieval-augmented generation methodology is its effectiveness in mitigating hallucinations caused by knowledge gaps, and their generality, which allows for application across any domain. This flexibility is further enhanced by the modularity of the approach, treating external knowledge bases like plug-ins that can be swapped or modified as needed.
In terms of the drawbacks, it can be easily impacted by irrelevant retrievals, which may decrease the overall performance by introducing noise or incorrect information into the response generation process.
Furthermore, the current paradigm exhibits shallow interactions between the retriever and generator components, leading to suboptimal knowledge utilization.
Hence, future research should focus on developing a robust RAG system that minimizes the impact of irrelevant retrieval, as well as integrating adaptive learning components that can dynamically adjust retrieval strategies based on the context of the query and the performance of previous interactions.

\subsection{Mitigating Training-related Hallucination}
\label{ssec:mitigating_training}
Training-related hallucinations typically arise from the intrinsic limitations of the architecture and training strategies adopted by LLMs. In this context, we discuss various optimization methods ranging from training stages (\S\ref{sssec:mitigating_training_pretraining}) and alignment stages (SFT \& RLHF) (\S\ref{sssec:mitigating_training_alignment}), aiming to mitigate hallucinations within the training process.

\subsubsection{Mitigating Pretraining-related Hallucination}
\label{sssec:mitigating_training_pretraining}
One significant avenue of research in mitigating pretraining-related hallucination centers on the limitations inherent in model architectures, especially \textit{unidirectional representation} and \textit{attention glitches}. In light of this, numerous studies have delved into designing novel model architectures specifically tailored to address these flaws.
To address the limitations inherent in unidirectional representation, \citet{li2023batgpt} introduced BATGPT which employs a bidirectional autoregressive approach. This design allows the model to predict the next token based on all previously seen tokens, considering both past and future contexts, thus capturing dependencies in both directions. Building on this idea, \citet{DBLP:journals/corr/abs-2307-03172} highlighted the potential of encoder-decoder models to make better use of their context windows, suggesting a promising direction for future LLMs architecture design.
Besides, recognizing the limitations of soft attention within self-attention-based architecture, \citet{liu2023exposing} proposed attention-sharpening regularizers. This plug-and-play approach specifies self-attention architectures using differentiable loss terms \citep{zhang2018attention} to promote sparsity, leading to a significant reduction in reasoning hallucinations. 

In the pre-training phase of LLMs, the choice of objective plays a pivotal role in determining the model's performance. However, conventional objectives can lead to fragmented representations and inconsistencies in model outputs. Recent advancements have sought to address these challenges by refining pre-training strategies, ensuring richer context comprehension, and circumventing biases. 
Addressing the inherent limitations in training LLMs, where unstructured factual knowledge at a document level often gets chunked due to GPU memory constraints and computational efficiency, leading to fragmented information and incorrect entity associations, \citet{lee2022factuality} introduced a factuality-enhanced training method. By appending a TOPICPREFIX to each sentence in factual documents, the approach transforms them into standalone facts, significantly reducing factual errors and enhancing the model's comprehension of factual associations. Similarly, considering that randomly concatenating shorter documents during pre-training might introduce inconsistencies in model outputs, \citet{shi2023context} proposed In-Context Pretraining, an innovative approach in which LLMs are trained on sequences of related documents. By altering the document order, this method aims to maximize similarity within the context windows. It explicitly encourages LLMs to reason across document boundaries, potentially bolstering the logical consistency between generations.

\paratitle{Discussion.}
Strategies designed to mitigate pretraining-related hallucinations typically are fundamental, potentially yielding significant improvements. However, they typically involve modifications to pre-training architectures and objectives, which are computationally intensive. Moreover, these integrations may lack broad applicability. Moving forward, the focus should be on developing adaptable and efficient strategies that can be universally applied without extensive system overhaul.
\subsubsection{Mitigating Misalignment Hallucination}
\label{sssec:mitigating_training_alignment}
Hallucinations induced during alignment often stem from capability misalignment and belief misalignment. However, defining the knowledge boundary of LLMs proves challenging, making it difficult to bridge the gap between LLMs' inherent capabilities and the knowledge presented in human-annotated data. While limited research addresses capability misalignment, the focus mainly shifts toward belief misalignment. 

Hallucinations stemming from belief misalignment often manifest as sycophancy, a tendency of LLMs to seek human approval in undesirable ways. This sycophantic behavior can be attributed to the fact that human preference judgments often favor sycophantic responses over more truthful ones \citep{sharma2023towards}, paving the way for reward hacking \citep{saunders2022self}. To address this, a straightforward strategy is to improve human preference judgments and, by extension, the preference model. Recent research \citep{bowman2022measuring, saunders2022self} has investigated the use of LLMs to assist human labelers in identifying overlooked flaws. Additionally, \citet{sharma2023towards} discovered that aggregating multiple human preferences enhances feedback quality, thereby reducing sycophancy. 

Besides, modifications to LLMs' internal activations have also shown the potential to alter model behavior. This can be achieved through methods like fine-tuning \citep{wei2023simple} or activation steering during inference \citep{dathathri2019plug, subramani2022extracting, hernandez2023inspecting}. Specifically, \citet{wei2023simple} proposed a synthetic-data intervention, fine-tuning language models using synthetic data where the claim's ground truth is independent of a user's opinion, aiming to reduce sycophantic tendencies. 

Another avenue of research \citep{rimsky2023reducing, rimsky2023modulating} has been to mitigate sycophancy through activation steering. This approach involves using pairs of sycophantic/non-sycophantic prompts to generate the sycophancy steering vector, derived from averaging the differences in intermediate activations. During inference, subtracting this vector can produce less sycophantic LLM outputs.

\paratitle{Discussion.}
Mitigating hallucinations through post-training methods represents a direct and effective approach, bypassing the complexities associated with data sourcing and pre-training.
However, a notable gap in current research is the limited attention given to capability misalignment within LLMs.
Future research should prioritize understanding the knowledge boundaries in capability alignment to address hallucinations effectively.

\subsection{Mitigating Inference-related Hallucination}
\label{ssec:mitigating_inference}
Decoding strategies in LLMs play a pivotal role in determining the factuality and faithfulness of the generated content. However, as analyzed in Section \S\ref{ssec:cause_decoding}, imperfect decoding often results in outputs that might lack factuality or stray from the original context. In this subsection, we explore two advanced strategies aimed at refining the decoding strategy to enhance both the factuality and faithfulness of the LLMs' outputs.

\subsubsection{Factuality Enhanced Decoding}

Factuality Enhanced Decoding aims to improve the reliability of outputs from LLMs by prioritizing the factuality of the information they generate. This line of methods focuses on aligning model outputs closely with established real-world facts, thereby minimizing the risk of disseminating false or misleading information.

\paratitle{Factuality Decoding.}
Considering the randomness in the sampling process can introduce non-factual content into open-ended text generation, \citet{lee2022factuality} introduced the factual-nucleus sampling algorithm that dynamically adjusts the nucleus probability $p$ throughout sentence generation. By dynamically adjusting the nucleus probability based on decay factors and lower boundaries and resetting the nucleus probability at the beginning of every new sentence, the decoding strategy strikes a balance between generating factual content and preserving output diversity. 
Moreover, some studies \citep{burns2022discovering, moschella2022relative} posit that the activation space of LLMs contains interpretable structures related to factuality. Building on this idea, \citet{li2023inference} introduced Inference-Time Intervention (ITI). This method first identifies a direction in the activation space associated with factually correct statements and then adjusts activations along the truth-correlated direction during inference. By repeatedly applying such intervention, LLMs can be steered towards producing more factual responses.
Similarly, \citet{chuang2023dola} delved into enhancing the factuality of LLM's decoding process from a perspective of factual knowledge storage. They exploit the hierarchical encoding of factual knowledge within transformer LLMs, noting that lower-level information is captured in earlier layers and semantic information in the later ones. Drawing inspiration from \cite{li2022contrastive}, they introduce DoLa, a strategy that dynamically selects and contrasts logits from different layers to refine decoding factuality. By placing emphasis on knowledge from higher layers and downplaying that from the lower layers, DoLa showcases its potential to make LLMs more factual, thus reducing hallucinations.

\paratitle{Post-editing Decoding.}
Unlike methods that directly modify the probability distribution to prevent hallucinations during the initial decoding, post-editing decoding seeks to harness the self-correction capabilities of LLMs \citep{pan2023automatically} to refine the originally generated content without relying on an external knowledge base. \citet{dhuliawala2023chain} introduced the Chain-of-Verification (COVE), which operates under the assumption that, when appropriately prompted, LLMs can self-correct their mistakes and provide more accurate facts. Starting with an initial draft, it first formulates verification questions and then systematically answers those questions in order to finally produce an improved revised response. 
Similarly, \citet{ji2023towards} focused on the medical domain and introduced an iterative self-reflection process. This process leverages the inherent ability of LLMs to first generate factual knowledge and then refine the response until it aligns consistently with the provided background knowledge.

\paratitle{Discussion.}
Factuality decoding methods, which typically assess the factuality at each decoding step, can offer substantial improvements.
Furthermore, due to their plug-and-play nature, they allow for application without the need for computation-intensive training.
Nevertheless, one of the primary limitations of these methods lies in balancing factual accuracy with maintaining the diversity and informativeness of the generated content, which can sometimes lead to compromises in either aspect.
On the other hand, post-editing decoding strategies, despite their effectiveness, heavily rely on the self-correction capabilities of LLMs, which may be unreliable.
Furthermore, applying self-reflection can be time-consuming, limiting their practicality for real-time applications.
Hence, it is crucial to achieve an optimal balance between factuality and computational efficiency.
\subsubsection{Faithfulness Enhanced Decoding}

On the other hand, Faithfulness Enhanced Decoding prioritizes alignment with the provided context and also emphasizes enhancing the consistency within the generated content.
Thus, in this section, we summarize existing work into two categories, including \textit{Context Consistency} and \textit{Logical Consistency}.

\paratitle{Context Consistency.}
In the era of LLMs, the issue of faithfulness hallucination typically lies in insufficient attention to the given context, which inspired numerous research to design inference-time strategies to enhance context consistency.
\citet{DBLP:journals/corr/abs-2305-14739} proposed context-aware decoding (CAD), which modifies the model's original output distribution in a contrastive formulation \citep{li2022contrastive}.
By amplifying the difference between output probabilities with and without context, CAD encourages the LLM to focus more on contextual information rather than over-rely on prior knowledge.
However, due to the inherent trade-off between diversity and context attribution \citep{zhang2020trading, gu2022improving}, overemphasizing contextual information can reduce diversity. 
To address this, \citet{DBLP:journals/corr/abs-2306-01286} introduced a dynamic decoding algorithm to bolster faithfulness while preserving diversity.
Specifically, the algorithm involves two parallel decoding steps, one with the context and one without. 
During the decoding, the KL divergence between two token distributions serves as a guiding signal, indicating the relevance of the source context.
This signal is utilized to dynamically adjust the sampling temperature to improve source attribution when the source is relevant.
In a parallel line of work, \citet{choi2023kcts} introduced knowledge-constrained decoding (KCD), which employed a token-level hallucination detection discriminator to identify contextual hallucinations and then guides the faithful generation process by reweighing the token distribution.
In addition to modifying output distribution in place to enhance contextual attention, another line of work has explored a generic post-edit approach to enhance faithfulness.
\citet{DBLP:conf/acl/GaoDPCCFZLLJG23} adopted a \textit{research-and-revise} workflow,  where the research stage raises questions about various aspects of the model's initial response and gathers evidence for each query, while the revision stage detects and revises any disagreements between the model's response and the evidence.
Similarly, \citet{lei2023chain} first detected contextual hallucinations at both the sentence and entity levels and then incorporated the judgments to refine the generated response.
Moreover, several studies have explored methods to overcome the softmax bottleneck, which constrains the expression of diversity and faithful representations. 
These approaches include employing a mixture of Softmax, which uses multiple hidden states to compute softmax multiple times and merge the resulting distributions \citep{yang2019mixtape} and incorporating pointer networks, which enables LLMs to copy the context words \citep{chang2023revisiting}, thereby reducing context hallucinations.

\paratitle{Logical Consistency.}
Inspired by the human thinking process, chain-of-thought \citep{wei2022chain} has been introduced to encourage LLMs to decompose complex problems into explicitly intermediate steps, thereby enhancing the reliability of the reasoning process \citep{chu2023a}.
Despite effective, recent research \citep{lanham2023measuring, turpin2023language} demonstrated that the intermediate rationales generated by LLMs do not faithfully capture their underlying behavior.
A branch of research has been inspired to improve the consistency of intermediate rationales generated by LLMs, particularly in multi-step reasoning \citep{cobbe2021training} and logical reasoning \citep{basso1993conditional}.
To enhance the self-consistency in chain-of-thought, \citet{wang2023scott} employed a knowledge distillation framework. They first generate a consistent rationale using contrastive decoding \citep{li2022contrastive} and then fine-tune the student model with a counterfactual reasoning objective, which effectively eliminates reasoning shortcuts \citep{branco2021shortcutted} that derive answers without considering the rationale. 
Furthermore, by employing contrastive decoding directly, LLMs can reduce surface-level copying and prevent missed reasoning steps \citep{o2023contrastive}.
In addition, \citet{li2024towards} conducted a deep analysis of the causal relevance among the context, CoT, and answer during unfaithful reasoning.
Analysis revealed that the unfaithfulness issue lies in the inconsistencies in the context information obtained by the CoT and the answer. 
To address this, they proposed inferential bridging, which takes the attribution method to recall contextual information as hints to enhance CoT reasoning and filter out noisy CoTs that have low semantic consistency and attribution scores to the context.
\citet{paul2024making} decomposed the reasoning process into two modules: an inference module, which employs Direct Preference Optimization \citep{rafailov2023direct} to align the LLM towards preferring correct reasoning chains over counterfactual chains, and a reasoning module, which encourages the LLM to reason faithfully over the reasoning steps using a counterfactual and causal preference objective.
Compared to natural language reasoning, logical reasoning demands rigorous logical calculation, whereas plain text often lacks precise logical structure, leading to unfaithful reasoning.
To address this, \citet{xu2024faithful} introduced Symbolic CoT (SymbCoT), which incorporates symbolic expressions within CoT to describe intermediate reasoning steps. Specifically, SymbCoT translates the natural language context into a symbolic representation and then formulates a step-by-step plan to address the logical reasoning problem, followed by a verifier to check the translation and reasoning chain, thereby ensuring faithful logical reasoning. 

\paratitle{Discussion.}
Faithfulness Enhanced Decoding significantly advances the alignment of LLM outputs with provided contexts and enhances the internal consistency of the generated content.
However, strategies such as context-aware decoding often lack adaptive mechanisms, limiting their effectiveness in scenarios that demand dynamic attention to context. Furthermore, many decoding strategies require the integration of additional models that do not focus on context, introducing significant computational overhead and reducing efficiency.

%% file: figures/retrieval.tex
\begin{figure*}[!t]
    \centering
    \includegraphics[width=0.8\textwidth]{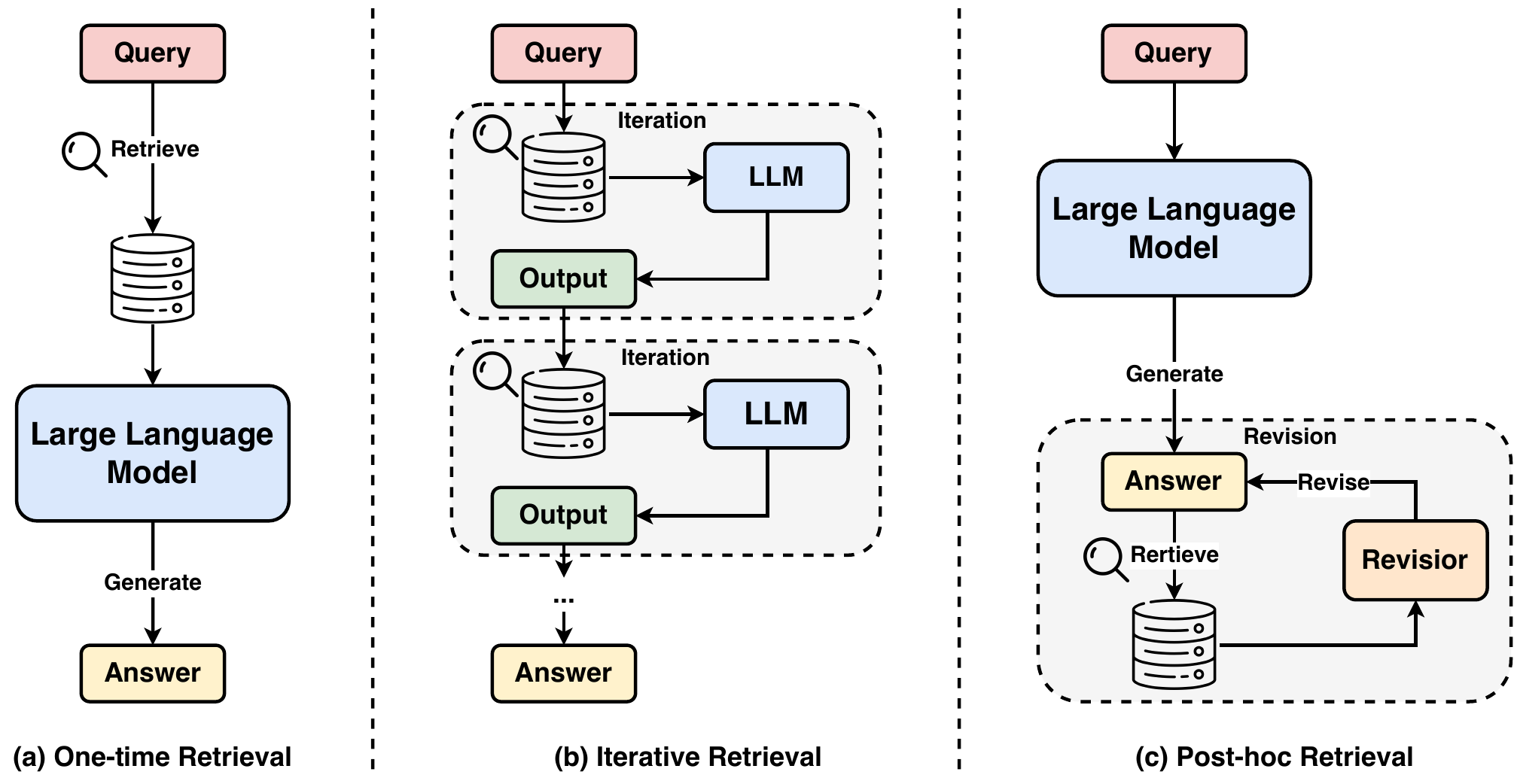}
    \caption{The illustration of three distinct approaches for Retrieval-Augmented Generation: \textbf{a) One-time Retrieval}, where relevant information is retrieved once before text generation; \textbf{b) Iterative Retrieval}, involving multiple retrieval iterations during text generation for dynamic information integration; and \textbf{c) Post-hoc Retrieval}, where the retrieval process happens after an answer is generated, aiming to refine and fact-check the generated content.}
    \label{fig:retrieval}
    \vspace{-0.4cm}
\end{figure*}

%% file: 6_RAG.tex
\section{Hallucinations in Retrieval Augmented Generation}
\label{sec:hallucination_rag}
Retrieval Augmented Generation (RAG) has emerged as a promising strategy to mitigate hallucinations and improve the factuality of LLM outputs \citep{lewis2020retrieval, izacard2023atlas, DBLP:journals/corr/abs-2302-00083, shi2023replug}.
By incorporating large-scale external knowledge bases during inference, RAG equips LLMs with up-to-date knowledge, thus reducing the potential risk of hallucination due to the inherent knowledge boundaries of LLMs \citep{DBLP:journals/corr/abs-2307-11019}.
Despite being designed to mitigate LLM hallucinations, retrieval-augmented LLMs can still produce hallucinations \citep{barnett2024seven}.
Hallucinations in RAG present considerable complexities, manifesting as outputs that are either factually inaccurate or misleading. These hallucinations occur when the content generated by the LLM does not align with real-world facts, fails to accurately reflect the user's query, or is not supported by the retrieved information.
Such hallucinations can stem from two primary factors: \textbf{retrieval failure} (\S\ref{ssec:rag_retrieval_failure}) and \textbf{generation bottleneck} (\S\ref{ssec:rag_generation_bottleneck}).
Through a comprehensive analysis of the limitations present in current RAG systems, we aim to shed light on potential improvements for retrieval-augmented LLMs, paving the way for more reliable information retrieval systems.

\subsection{Retrieval Failure}
\label{ssec:rag_retrieval_failure}
The retrieval process is a crucial initial step in the RAG framework, tasked with retrieving the most relevant information for information-seeking queries.
Consequently, failures in the retrieval stage can have serious downstream effects on the RAG pipeline, leading to hallucinations. These failures typically stem from three primary parts: the formulation of user queries, the reliability and scope of retrieval sources, and the effectiveness of the retriever.

\subsubsection{User Queries}
User queries play a fundamental role in guiding the retrieval process with RAG systems.
The specificity and clarity of these queries critically influence the effectiveness of retrieval outcomes.
In this section, we discuss factors that may contribute to hallucinations from three perspectives: blind retrieval, misinterpretation of ambiguous queries, and the challenges in accurate retrieval of complex queries. Some examples are presented in Table \ref{tab:user_query_example} for a better understanding.
\input{tables/rag_user_queries}

\paratitle{Retrieval Intent Decisions.}
Not all queries necessitate retrieval.
Blind retrieval for queries that do not require external knowledge can counterproductively lead to misleading responses.
As shown in Table \ref{tab:user_query_example}, the query about \textit{"the boiling point of water at sea level"} pertains to a basic scientific fact that the model could address without external retrieval. However, the retrieval system was inappropriately activated, blindly retrieving inaccurate information and consequently leading to an undesirable response.
Consequently, several studies \citep{DBLP:conf/acl/MallenAZDKH23, ding2024retrieve, ni2024when, zhang2024retrievalqa} have proposed to make a shift from passive retrieval to adaptive retrieval.
In general, these strategies can be divided into two categories: \textit{heuristic-based} and \textit{self-aware judgment}.
\textit{Heuristic-based} methods employ heuristic rules to determine the necessity of retrieval.
For instance, \citet{DBLP:conf/acl/MallenAZDKH23} observed a positive correlation between LLMs' memorization capabilities and entity popularity and suggested triggering retrieval only when the entity popularity in the user query falls below a certain threshold.
Similarly, \citet{jeong2024adaptive} determined the timing of retrieval based on the query complexity, whereas \citet{asai2023selfrag} considered whether the query is factual relevant.
\textit{Self-aware judgment} leverages the models' intrinsic judgment to decide the necessity for information retrieval.
\citet{feng2023cook}, \citet{ren2023investigating} and \citet{wang2023self} directly prompted LLMs for retrieval decisions, recognizing that LLMs possess a certain level of awareness regarding their knowledge boundaries \citep{kadavath2022language, yin2023do}.
Moreover, \citet{DBLP:journals/corr/abs-2305-06983} introduced an active retrieval strategy that triggers retrieval only when the LLM generates low-probability tokens.
Similarly, \citet{su2024dragin} not only considered the uncertainty of each token but also its semantic contribution and impact on the subsequent context.
More recently, \citet{cheng2024unified} proposed four orthogonal criteria for determining the retrieval timing, which include intent-aware, knowledge-aware, time-sensitive-aware, and self-aware.

\paratitle{Ambiguous Queries.}
Ambiguous user queries, containing omission, coreference, and ambiguity, significantly complicate the retrieval system's ability to fetch precisely relevant information, thereby increasing the likelihood of generating undesirable responses.
As shown in Table \ref{tab:user_query_example}, due to the ambiguity of the query about \textit{"the record for the fastest mile run on track"}, the retrieval system erroneously retrieved information from automobile racing events, which led the model to generate a response suited for vehicles instead of athletes.
A prevalent mitigation strategy is query rewriting, where queries are refined and decontextualized to better match relevant documents.
\citet{DBLP:conf/emnlp/WangYW23} and \citet{jagerman2023query} have explored prompting approaches where the LLM is prompted to generate a pseudo-document or rationale based on the original query, which is then used for further retrieval.
Additionally, \citet{ma2023query} introduced a trainable rewriter which is trained using the feedback from the LLM via reinforcement learning.
\citet{mao2024rafe} employed the feedback signals from the reranker to train the rewrite model, thus eliminating the reliance on annotated data. 
However, the challenges deepen in conversational search, which encounters a more complex issue of context-dependent query understanding with the lengthy conversational history.
Addressing this, \citet{DBLP:journals/corr/abs-2402-11827} proposed a similar framework for optimizing the LLM to generate retriever-preferred query rewrites. This operated by generating a variety of queries and then using the preference of the rank of retrieved passage to optimize the query rewriting model.

\paratitle{Complex Queries.}
Complex user queries, characterized by requiring intensive reasoning \citep{su2024brightrealisticchallengingbenchmark} or encompassing multiple aspects \citep{shao2024assisting, wang2024richrag}, pose significant challenges to the retrieval system. 
Such queries require advanced understanding and decomposition capabilities, which may exceed the current capabilities of the current retrieval methods based on keyword or semantic matching, often leading to partial or incorrect retrievals.
For example, as shown in Table \ref{tab:user_query_example}, due to the multi-step nature of the query about \textit{"Which university did the scientist who discovered penicillin graduate from?"}, direct retrieval often leads to incomplete results, thereby resulting in hallucinatory responses.
A common approach involves query decomposition, where the complex query is decomposed into sub-queries to facilitate more accurate information retrieval.
For instance, \citet{wang2024richrag} implemented a sub-aspect explorer that utilizes the extensive world knowledge embedded LLMs to identify potential sub-aspects of user queries, thereby providing explicit insights into the user's underlying intents.
Similarly, \citet{shao2024assisting} concentrated on the demanding task of expository writing, aiming at retrieving comprehensive information to compose Wikipedia-like articles from scratch on a specific topic.
This approach involves decomposing the topic into various perspectives and simulating multi-turn conversations with LLMs, each personified with different perspectives for question asking.
Additionally, \citet{DBLP:conf/emnlp/CaoZSL0THL23} and \citet{chu2024beamaggrbeamaggregationreasoning} explored knowledge-intensive complex reasoning and employed a divide-and-conquer strategy. This strategy begins with decomposing complex questions into question trees, where at each node, the LLM retrieves and aggregates answers from diverse knowledge sources.

\subsubsection{Retrieval Sources}
The reliability and scope of retrieval sources are crucial determinants of the efficacy of RAG systems. Effective retrieval depends not only on the clarity of the user queries but also on the quality and comprehensiveness of the sources from which information is retrieved. When these sources contain factually incorrect or outdated information, the risk of retrieval failures increases significantly, potentially leading to the generation of incorrect or misleading information.

As the landscape of content creation evolves with the rapid advancement of Artificial Intelligence Generated Content (AIGC) \citep{cao2023a}, an increasing volume of LLM-generated content is permeating the internet, subsequently becoming integrated into retrieval sources \citep{chen2023combating}. This integration is reshaping the dynamics of information retrieval, as evidenced by recent empirical studies \citep{dai2023llms, xu2023ai} suggesting that modern retrieval models tend to favor LLM-generated content over human-authored content. 
Recent research \citep{chen2024spiralsilencelargelanguage} has explored the implications of progressively integrating LLM-generated content into RAG systems. The findings indicate that, without appropriate intervention, human-generated content may progressively lose its influence within RAG systems.
Additionally, \citet{tan2024blinded} investigated the performance of RAG systems when incorporating LLM-generated into retrieved contexts, revealing a significant bias favoring generated contexts.
This bias stems from the high similarity between generated context and questions, as well as the semantic incompleteness of retrieved contexts.
More seriously, the propensity of LLMs to produce factually inaccurate hallucinations exacerbates the reliability issues of retrieval sources. 
As LLM-generated content often contains factual errors, its integration into retrieval sources can mislead retrieval systems, further diminishing the accuracy and reliability of the information retrieved.

To combat these biases, several approaches have been explored.
Inspired by common practice in pre-training data processing \citep{black2022gpt}, \citet{asai2024reliable} proposed a scenario that incorporates a quality filter designed to ensure the high quality of the retrieval datastore.
Additionally, \citet{pan2024not} proposed Credibility-aware Generation (CAG), which equips LLMs with the ability to discern and handle information based on its credibility.
This approach assigns different credibility levels to information, considering its relevance, temporal context, and the trustworthiness of its source, thus effectively reducing the impact of flawed information in RAG systems.

\subsubsection{Retriever}
When the user query is explicit and the retrieval source is reliable, the effectiveness of the retrieval process depends crucially on the performance of the retriever.
In such scenarios, the retriever's effectiveness is significantly compromised by improper chunking and embedding practices.

\paratitle{Chunking.}
Given the extensive nature of retrieval sources, which often encompass lengthy documents like web pages, it poses significant challenges for LLMs with limited context length.
Thus, chunking emerges as an indispensable step in RAG, which involves segmenting these voluminous documents into smaller, more manageable chunks to provide precise and relevant evidence for LLMs.
According to actual needs, the chunking granularity ranges from documents to paragraphs, even sentences.
However, inappropriate retrieval granularity can compromise the semantic integrity and affect the relevance of retrieved information \citep{nair2023a}, thereby affecting the performance of LLMs.
Fixed-size chunking, which typically breaks down the documents into chunks of a specified length such as 100-word paragraphs, serves as the most crude and prevalent strategy of chunking, which is widely used in RAG systems \citep{guu2020retrieval, lewis2020retrieval, borgeaud2022improving}.
Considering fixed-size chunking falls short in capture structure and dependency of lengthy documents, \citet{sarthi2024raptor} proposed RAPTOR, an indexing and retrieval system. By recursively embedding, clustering, and summarizing chunks of text, RAPTOR constructs a tree to capture both high-level and low-level details.
When retrieval, RAPTOR enables LLMs to integrate information from different levels of abstraction, providing a more comprehensive context for user queries.
Instead of chunking text with a fixed chunk size, semantic chunking adaptively identifies breakpoints between sentences through embedding similarity, thereby preserving semantic continuity \citep{kamradt2024the}.
Furthermore, \citet{chen2023dense} pointed out the limitations of the existing retrieval granularity.  
On the one hand, while a coarser retrieval with a longer context can theoretically provide a more comprehensive context, it often includes extraneous details that could potentially distract LLMs. On the other hand, a fine-grain level can provide more precise and relevant information, it has limitations such as not being self-contained and lacking necessary contextual information.
To address these shortcomings, \citet{chen2023dense} introduced a novel retrieval granularity, proposition, which is defined as atomic expressions within the text, each encapsulating a distinct factoid and presented in a concise self-contained natural language format.

\paratitle{Embedding.}
Once the retrieval text is chunked, text chunks are subsequently transformed into vector representation via an embedding model.
Such a representation scheme is supported by the well-known data structure of \textit{vector database} \citep{jing2024when}, which systematically organizes data as key-value pairs for efficient text retrieval.
In this manner, the relevance score can be computed according to the similarity function between the text representation and query representation.
However, a sub-optimal embedding model may compromise performance, which affects the similarity and matching of chunks to user queries, potentially misleading LLMs.
Typically, a standard embedding model \citep{izacard2022unsuperised, gao2021simcse, karpukhin2020dense, zhao2024dense} learns the query and text representations with encoder-based architecture (\textit{e.g.} BERT \citep{devlin2019bert}, RoBERTa \citep{liu2020roberta}) via contrastive learning \citep{oord2018representation}, where the loss is constructed by contrasting a positive pair of query-document against a set of random negative pairs.
However, these embeddings showcase their limitations when applied to new domains, such as medical and financial applications \citep{thakur2021beir, muennighoff2023mteb}.
In these cases, recent studies \citep{shi2023replug, xiao2023cpack, ovadia2023finetuning, balaguer2024rag} propose to fine-tune the embedding models on domain-specific data to enhance retrieval relevance.
For example, REPLUG \citep{shi2023replug} utilizes language modeling scores of the answers as a proxy signal to train the dense retriever.
More recently, \citet{muennighoff2024generative} have introduced generative representational instruction tuning where a single LLM is trained to handle both generative and embedding tasks, which largely reduces inference latency in RAG by caching representations.
Despite these advancements, the field faces challenges, particularly with the fine-tuning of high-performing yet inaccessible embedding models, such as OpenAI's text-embedding-ada-002.
Addressing this gap, \citet{zhang2024mafin} introduced a novel approach for fine-tuning a black-box embedding model by augmenting it with a trainable embedding model which significantly enhances the performance of the black-box embeddings.

\subsection{Generation Bottleneck}
\label{ssec:rag_generation_bottleneck}
After the retrieval process, the generation stage emerges as a pivotal point, responsible for generating content that faithfully reflects the retrieved information. However, this stage can encounter significant bottlenecks that may lead to hallucinations. We summarize two key capabilities of LLMs that are closely related to these bottlenecks: contextual awareness and contextual alignment. Each plays an important role in ensuring the reliability and credibility of the RAG system.
\subsubsection{Contextual Awareness}
\label{sssec:contexual_awareness}
Contextual awareness involves understanding and effectively utilizing contextual information retrieved. This section discusses the key factors that impact the LLM's ability to maintain contextual awareness, which can be categorized into three main parts: (1) the presence of noisy retrieval in context, (2) context conflicts, and (3) insufficient utilization of context information.

\paratitle{Noisy Context.}
As emphasized in \S\ref{ssec:rag_retrieval_failure}, the failure in the retrieval process may inevitably introduce irrelevant information, which will propagate into the generation stage.
When the generator is not robust enough to these irrelevant retrievals, it will mislead the generator and even introduce hallucinations \citep{cuconasu2024the}.

\citet{yoran2023making} conducted a comprehensive analysis on the robustness of current retrieval-augmented LLMs, revealing a significant decrease in performance with random retrieval. While using an NLI model to filter out irrelevant passages is effective, this method comes with the trade-off of inadvertently discarding some relevant passages. A more effective solution is to train LLMs to ignore irrelevant contexts by incorporating irrelevant contexts in training data.
Similarly, \citet{yu2023chain} introduced Chain-of-Note, which enables LLMs to first generate reading notes for retrieved contexts and subsequently formulate the final answer. In this way, LLMs can not only filter irrelevant retrieval to improve noise robustness but also respond with unknown when retrieval is insufficient to answer user queries. 
In addition to improving LLM robustness by learning to ignore irrelevant content in the context, several studies \citep{li2023unlocking, jiang2023llmlingua, xu2023recomp, wang2023learning} propose to compress the context to filter out irrelevant information.
Specifically, \citet{li2023unlocking} and \citet{jiang2023llmlingua} made use of small language models to compute self-information and perplexity for prompt compression, finding the most informative content. Similarly, \citet{wang2023learning} proposed to filter out irrelevant content and leave precisely supporting content based on lexical and information-theoretic approaches. 
Besides, efforts have been also made to employ summarization models as compressors. \citet{xu2023recomp} presented both extractive and abstractive compressors, which are trained to improve LLMs' performance while keeping the prompt concise. \citet{liu2023tcra} involved summarization compression and semantic compression, where the former achieves compression by summarizing while the latter removes tokens with a lower impact on the semantic.

\paratitle{Context Conflict.}
Retrieval-augmented LLMs generate answers through the combined effect of parametric knowledge and contextual knowledge. As discussed in \S\ref{sssec:cause_over_confidence}, LLMs may sometimes exhibit over-confidence, which can bring new challenges to the faithfulness of RAG systems when facing knowledge conflicts.
Knowledge conflicts in RAG are situations where contextual knowledge contradicts LLMs' parametric knowledge.
\citet{longpre2021entity} first investigated knowledge conflicts in open-domain question answering, where conflicts are automatically created by replacing all spans of the gold answer in the retrieval context with a substituted entity.
Findings demonstrate that generative QA reader models (\textit{e.g.} T5) tend to trust parametric memory over contextual information. By further training the retriever to learn to trust the contextual evidence with augmented training examples by entity substitution, the issue of over-reliance on parametric knowledge is mitigated.
Similar findings are also reported by \citet{DBLP:conf/acl/LiRZWLVYK23} who demonstrated that fine-tuning LLMs on counterfactual contexts can effectively improve the controllability of LLMs when dealing with contradicts contexts.
Also building upon counterfactual data augmentation, \citet{neeman2023disentqa} trained models to predict two disentangled answers, one based on contextual knowledge and the other leveraging parametric knowledge to address knowledge conflicts.
Besides, \citet{zhou2023context} introduced two effective prompting-based strategies, namely opinion-based prompts and counterfactual demonstrations.
Opinion-based prompts transform the context to narrators' statements, soliciting the narrators' opinions, whereas counterfactual demonstrations employ counterfactual instances to improve faithfulness in situations of knowledge conflict.
While \citet{longpre2021entity} and \citet{DBLP:conf/acl/LiRZWLVYK23} concentrated their research on the context of a limited single evidence setting, \citet{chen2022rich} further expanded this study to consider a more realistic scenario in which models consider multiple evidence passages and find models rely almost exclusively on contextual evidence.

Considering previous studies \citep{longpre2021entity, DBLP:conf/acl/LiRZWLVYK23} mostly focused on smaller models, \citet{DBLP:journals/corr/abs-2305-13300} raised doubts about the applicability of their conclusions in the era of LLMs. Such heuristic entity-level substitution may lead to incoherent counter-memory, thereby making it trivial for LLMs to overlook the construct knowledge conflicts.
By directly eliciting LLMs to generate a coherent counter-memory that factually conflicts with the parametric memory, LLMs exhibit their high receptivity to external evidence.

\paratitle{Context Utilization.}
Despite successfully retrieving evidence relevant to factoid queries, LLMs can encounter a significant performance degradation due to insufficient utilization of the context, especially for information located in the middle of the long context window, a notable issue known as the \textit{lost-in-the-middle} phenomenon \citep{DBLP:journals/corr/abs-2307-03172}.
Beyond factoid QA, recent studies have further demonstrated such a \textit{middle-curse} also holds in abstractive summarization \citep{ravaut2024context}, long-form QA \citep{chen2023understanding} and passage ranking \citep{tang2023found}. One potential explanation lies in the use of rotary positional embedding (RoPE) \citep{su2021roformer}, which is widely used in open-source LLMs, due to its excellent performance in length extrapolation \citep{zhao2023length}. As a representative relative position embedding, RoPE features a long-term decay property, which inherently biases the LLM to give precedence to current or proximate tokens, thereby diminishing its attention on those that are more distant.
Another contributing factor is that the most salient information often resides at the beginning or the end of pre-training data, a characteristic commonly observed in news reports \citep{ravaut2024context}.
Such an issue brings forth challenges in retrieval-augmented LLMs, as retrieval-augmented LLMs are typically designed with extensive lengths to accommodate more retrieval documents.

To mitigate this crucial issue, \citet{he2023never} introduced several tasks specially designed for information seeking to enhance the capability of information utilization by explicitly repeating the question and extracting the index of supporting documents before generating answers.
Furthermore, \citet{zhang2024middle} introduced Multi-scale Positional Encoding (Ms-PoE), which mitigates the long-term decay effect characteristic of RoPE by rescaling position indices. Ms-PoE provides a plug-and-play solution to enhance the ability of LLMs to effectively capture information in the middle of the context without the need for additional fine-tuning.
Besides, \citet{ravaut2024context} proposed hierarchical and incremental summarization, which effectively preserves the salient information and compresses the length of context to avoid the \textit{middle-curse}.

\subsubsection{Contextual Alignment}
Contextual alignment ensures that LLM outputs faithfully align with relevant context. This section outlines the primary components of contextual alignment, which include: (1) source attribution and (2) faithful decoding.

\paratitle{Source Attribution.}
Source attribution \citep{huang2024advancing} in retrieval-augmented LLMs refers to the process by which the model identifies and utilizes the origins of information within its generation process. This component is crucial for ensuring that the outputs of RAG systems are not only relevant but also verifiable and grounded in credible sources.

To achieve source attribution in RAG systems, recent studies have been explored, which can be categorized into three lines based on the type of attribution.
(1) \textit{Plan-then-Generate}: \citet{fierro2024learning} introduced the blueprint model for attribution, which conceptualizes text plans as a series of questions that serve as blueprints for generation process, dictating both the content and the sequence of the output. Compared with abstractive questions, \citet{huang2024learning} enabled the model to first ground to extractive evidence spans, which guides the subsequent generation process. Leveraging either abstract questions or extractive spans as planning facilitates a built-in attribution mechanism, as they provide a natural link between retrieved information and the subsequent generation. Similarly, \citet{slobodkin2024attribute} broke down the conventional end-to-end generation process into three intuitive stages: content selection, sentence planning, and sentence fusion. By initially identifying relevant source segments and subsequently conditioning the generation process on them, the selected segments naturally serve as attributions. (2) \textit{Generate-then-Reflect}: \citet{asai2023selfrag} proposed training the LLM to generate text with reflection tokens. These reflection tokens empower the LLM to decide whether to retrieve, assess the relevance of the retrieved document, and critique its own generation to ensure attributability. By critiquing its generation. Furthermore, \citet{ye2023effective} introduced AGREE, designed to facilitate self-grounding in LLMs. AGREE trains LLMs to generate well-grounded claims with citations and identify claims that lack verification. An iterative retrieval process is then employed to actively seek additional information for these unsupported statements.
(3) \textit{Self-Attribution}: In addition to leveraging external supervised signals for attribution, \citet{qi2024model} proposed a self-attribution mechanism that utilizes model-internal signals. It operates by first identifying context-sensitive answer tokens, which are then paired with retrieved documents that contributed to the model generation via saliency methods.

\paratitle{Faithful Decoding.}
Despite significant optimizations in the RAG pipeline that facilitate the incorporation of highly relevant content into the model's context, current LLMs still cannot guarantee faithful generation. The unfaithful utilization of relevant context by LLMs undermines the reliability of their outputs, even when the sources of information are verifiably accurate. 
\citet{wu2024clashevalquantifyingtugofwarllms} analyzed the model's knowledge preference when internal knowledge conflicts with contextual information and observed the tug-of-war between the LLM’s internal prior and external evidence. 
To tackle this issue, recent research \citep{DBLP:journals/corr/abs-2305-14739, wu2024synchronous} has focused on faithful decoding within RAG systems, aiming to improve the models' ability to generate content that faithfully aligns with contextual information.
\citet{DBLP:journals/corr/abs-2305-14739} presented context-aware decoding, which modifies the model's original output probability distribution into the pointwise mutual information (PMI) formulation. The strategy operates by amplifying the difference between the output probabilities when a model is used with and without context, thereby enhancing the faithfulness of LLMs to the provided context.
\citet{li2024nearest} adopted a semi-parametric language modeling approach \citep{khandelwal2020generalization} which facilitates the integration of contextual spans of arbitrary length into LM generations. The generation is then verified via speculative decoding, further ensuring model faithfulness.
More recently, \citet{wu2024synchronous} proposed faithfulness-oriented decoding, which leverages a lightweight faithfulness detector to monitor the beam-search process. The detector leverages fine-grained decoding dynamics including sequence likelihood, uncertainty quantification, context influence, and semantic alignment to synchronously detect unfaithful sentences. When an unfaithful generation is detected, it triggers the backtrack operation and selects the beam with the more faithful score, thus ensuring greater faithfulness to the retrieval sources.

%% file: tables/rag_user_queries.tex
\begin{table*}[]
\centering
\caption{Examples of each category of User Queries that may lead to LLM hallucinations. Content marked in \textcolor{red}{Red} represents the hallucinatory output.}
  \label{tab:user_query_example}
\vspace{-4mm}  
\begin{adjustbox}{max width=\textwidth}\setlength{\tabcolsep}{0.3mm}
{
\begin{tabular}{lll}
\toprule
\textbf{Type} & \textbf{Example Dialog} & \textbf{Explanation} 
\\\midrule
\begin{tabular}[c]{@{}l@{}} Retrieval \\ Intent \end{tabular} &
  \begin{tabular}[c]{@{}l@{}}
  \textbf{Retrieved context:} At high altitudes, water boils at lower\\ temperatures than at sea level, such as 95°C at elevations \\above 2,000 meters. This altitude-induced reduction can affect \\cooking times and methods. \\
   \midrule
  \textbf{Query:} What is the boiling point of water at sea level? \\\midrule  \textbf{Response:} The boiling point of water at sea level is \textcolor{red}{95°C}.
  \\
  \end{tabular} &
  \begin{tabular}[c]{@{}l@{}}
    \textbf{This query involves a basic scientific fact that should be} \\ \textbf{available within the model's parametric knowledge.}\\ \textbf{However, the system inappropriately retrieves} and uses \\altitude-specific information about boiling temperatures, \\leading to an incorrect response for sea-level conditions.
    \end{tabular}
   \\ \cmidrule(lr){1-3}
  \begin{tabular}[c]{@{}l@{}} Ambiguous \\ Queries \end{tabular} &
  \begin{tabular}[c]{@{}l@{}}
  \textbf{Retrieved context:} The Hennessey Venom GT famously \\set a speed record by completing a mile on a racetrack in \\32 seconds, highlighting its performance in motorsport events. \\
   \midrule
  \textbf{Query:} What's the record for the fastest mile run on track? \\\midrule  \textbf{Response:} The fastest mile on a track was completed in \\ \textcolor{red}{32 seconds by the Hennessey Venom GT.}
  \\
  \end{tabular} &
  \begin{tabular}[c]{@{}l@{}}
    \textbf{The query is ambiguous because it does not specify whe-} \\ \textbf{ther it refers to a human runner or a vehicle.} The retrieval \\ system erroneously pulls information related to car racing \\ records instead of athletics.
    \end{tabular}
   \\ \cmidrule(lr){1-3}
\begin{tabular}[c]{@{}l@{}} Complex \\ Queries \end{tabular} &
  \begin{tabular}[c]{@{}l@{}}
  \textbf{Retrieved context:} Alexander Fleming, known for his ground-\\ breaking work in discovering penicillin, significantly influenced \\ modern antibiotics. Fleming's discovery saved countless lives. \\
   \midrule
  \textbf{Query:} Which university did the scientist who discovered \\ penicillin graduate from?
  \\
  \midrule  
  \textbf{Response:} \textcolor{red}{Alexander Fleming graduated from Oxford University.}
  \end{tabular} &
  \begin{tabular}[c]{@{}l@{}}
    \textbf{The retrieval system only manages to fetch information} \\ \textbf{about Fleming's professional achievements in the discovery} \\ \textbf{of penicillin}. However, the document does not provide informa- \\ tion about his educational background, thus the model generates a \\ hallucinatory answer.
    \end{tabular}
  \\\bottomrule
\end{tabular}
}
\end{adjustbox}
\end{table*}

%% file: 7_Future.tex
\section{Future discussion}
\label{sec:future}
As the field of research on hallucinations in LLMs continues to evolve, our focus shifts towards the next horizon of inquiry. We explore prospective areas of study, notably the phenomenon of hallucinations in vision-language models (\S\ref{ssec:hallucination_lvlm}) and the challenge of delineating and understanding knowledge boundaries within LLMs (\S\ref{ssec:hallucination_knowledge_boundary}). 

\subsection{Hallucination in Large Vision-Language Models}
\label{ssec:hallucination_lvlm}
Enabling the visual perception ability, along with exceptional language understanding and generation capabilities, Large Vision-Language Models (LVLMs) have exhibited remarkable vision-language capabilities \citep{zhu2023minigpt, liu2023visual, yu2021hybrid, huang2023language, maaz2023video, chen2023measuring, yu2023knowledge, zellers2019recognition}. Unlike previous pre-trained multi-modal models that gain limited vision-language abilities from large-scale visual-language pre-training datasets \citep{wang2021simvlm, li2023blip, luo2020univl, zhong2023stoavlp}, LVLMs exploit advanced LLMs to unleash the power of interacting with humans and the environment. The consequent diverse applications of LVLMs also bring new challenges to maintaining the reliability of such systems. Recent studies have revealed that current LVLMs are suffering from multi-modal hallucinations, where models provide responses misaligned with the corresponding visual information \cite{liu2024survey, tong2024eyes,guan2023hallusionbench}. Such multi-modal hallucinations could cause unexpected behaviors when applying LVLMs to real-world scenarios, which therefore had to be further investigated and mitigated. 

\citet{li2023evaluating} and \citet{lovenia2023negative} took the first step towards evaluating the object hallucinations in the LVLMs. Evaluations and experiments reveal that current LVLMs are prone to generate inconsistent responses with respect to the associated image, including non-existent objects, wrong object types, and attributes, incorrect semantic relationships, etc. \cite{wang2023llm, zhai2023halle}. Furthermore, \citet{liu2023mitigating}, \citet{zong2023fool} and \citet{liu2023hallusionbench} show that LVLMs can be easily fooled and experience a severe performance drop due to their over-reliance on the strong language prior, as well as its inferior ability to defend against inappropriate user inputs \cite{jeong2023hijacking,han2024instinctive}. \citet{jiang2024hal}, \citet{wang2023llm} and \citet{jing2023faithscore} took a step forward to holistically evaluate multi-modal hallucination. What's more, when presented with multiple images, LVLMs sometimes mix or miss parts of the visual context, as well as fail to understand temporal or logical connections between them, which might hinder their usage in many scenarios, yet properly identifying the reason for such disorders and tackling them still requires continued efforts. Despite the witnessed perception errors, LVLMs can generate flawed logical reasoning results even when correctly recognizing all visual elements, which remains further investigation.

Efforts have been made towards building a more robust large vision-language model. \citet{gunjal2023detecting}, \citet{lu2023evaluation}, \citet{wang2024mitigating}, and \citet{liu2023mitigating} proposed to further finetune the model for producing more truthful and helpful responses. Another line of work chooses to post-hoc rectify the generated inconsistent content, such as \cite{zhou2023analyzing}, and \cite{yin2023woodpecker}, which introduced expert models. To free from the external tools, \citet{leng2023mitigating}, \citet{huang2023opera}, and \citet{zhao2024mitigating} tried to fully utilize the LVLM itself to alleviate hallucinations. Though proved to be effective, those methods usually require additional data annotations, visual experts, training phases, and computational costs, which prevent LVLMs from effectively scaling and generalizing to various fields. Thus, more universal approaches are expected to build a more reliable system, such as faithful and large-scale visual-text pre-training and alignment methods.

\subsection{Understanding Knowledge Boundary in LLMs}
\label{ssec:hallucination_knowledge_boundary}

Despite the impressive capacity to capture factual knowledge from extensive data, LLMs still face challenges in recognizing their own knowledge boundaries. This shortfall leads to the occurrence of hallucinations, where LLMs confidently produce falsehoods without an awareness of their own knowledge limits \citep{pacchiardi2023catch,ren2023investigating, zhao2023knowing}. Numerous studies delve into probing knowledge boundaries of LLMs, utilizing strategies such as evaluating the probability of a correct response in a multiple-choice setting \citep{kadavath2022language}, or quantifying the model's output uncertainty by evaluating the similarity among sets of sentences with uncertain meanings. 

Furthermore, a line of work \citep{moschella2022relative, burns2022discovering, li2023inference, DBLP:journals/corr/abs-2304-13734} has revealed that LLMs contain latent structures within their activation space that relate to beliefs about truthfulness. Recent research \citep{slobodkin2023curious} also found substantial evidence for LLMs' ability to encode the unanswerability of questions, despite the fact that these models exhibit overconfidence and produce hallucinations when presented with unanswerable questions. Nonetheless, \citet{levinstein2023still} have employed empirical and conceptual tools to probe whether or not LLMs have beliefs. Their empirical results suggest that current lie-detector methods for LLMs are not yet fully reliable, and the probing methods proposed by \citet{burns2022discovering} and \citet{DBLP:journals/corr/abs-2304-13734} do not adequately generalize. Consequently, whether we can effectively probe LLMs' internal beliefs is ongoing, requiring further research.

%% file: 8_Conclusion.tex
\section{Conclusion}
\label{sec:conclusion}

In this comprehensive survey, we have undertaken an in-depth examination of hallucinations within large language models, delving into the intricacies of their underlying causes, pioneering detection methodologies as well as related benchmarks, and effective mitigation strategies. Although significant strides have been taken, the conundrum of hallucination in LLMs remains a compelling and ongoing concern that demands continuous investigation. Moreover, we envision this survey as a guiding beacon for researchers dedicated to advancing robust information retrieval systems and trustworthy artificial intelligence. By navigating the complex landscape of hallucinations, we hope to empower these dedicated individuals with invaluable insights that drive the evolution of AI technologies toward greater reliability and safety.